\documentclass[letterpaper]{article}
\usepackage[preprint]{aaai2027}
\usepackage[hyphens]{url}
\usepackage{graphicx}
\usepackage{natbib}
\usepackage{caption}
\usepackage{amsmath}
\usepackage{amssymb}
\usepackage{pifont}
\usepackage{booktabs}
\usepackage{array}
\usepackage{algorithm}
\usepackage{algorithmic}

\newcommand{\method}{DUET}
\newcommand{\methodplus}{\method{}+}
\newcommand{\duetgradient}{%
  \textcolor[HTML]{4472C4}{D}%
  \textcolor[HTML]{7B7286}{U}%
  \textcolor[HTML]{B17148}{E}%
  \textcolor[HTML]{E8710A}{T}%
}
\newcommand{\duetplusgradient}{\duetgradient\textcolor[HTML]{E8710A}{+}}
\usepackage{colortbl}
\newcommand{\first}[1]{\cellcolor{red!35}#1}
\newcommand{\second}[1]{\cellcolor{blue!25}#1}
\newcommand{\third}[1]{\cellcolor{green!30}#1}
\newcommand{\scmexpert}{sCM expert}
\newcommand{\dmdexpert}{DMD expert}

\newcommand{\nfes}{NFE}
\newcommand{\pdata}{p_{\mathrm{data}}}
\newcommand{\Normal}{\mathcal{N}}
\newcommand{\E}{\mathbb{E}}

\title{\duetgradient: A Diversity--Quality Duet of Distillation Experts for Two-Step Video Generation}
\author{
    Zian Li\textsuperscript{\rm 1,2}\thanks{Work done as an intern at Alibaba Group (zian@stu.pku.edu.cn).},\hspace{0.5em}
    Litong Gong\textsuperscript{\rm 4},\hspace{0.5em}
    Borui Liao\textsuperscript{\rm 4},\hspace{0.5em}
    Pengfei Liu\textsuperscript{\rm 4},\hspace{0.5em}
    Xinyu Wang\textsuperscript{\rm 5},\\
    Xinyuan Wei\textsuperscript{\rm 1},\hspace{0.5em}
    Yifan Gao\textsuperscript{\rm 4},\hspace{0.5em}
    Tiezheng Ge\textsuperscript{\rm 4},\hspace{0.5em}
    Muhan Zhang\textsuperscript{\rm 1,3}\thanks{Correspondence to Muhan Zhang (muhan@pku.edu.cn).}
}
\affiliations{
    \textsuperscript{\rm 1}Institute for Artificial Intelligence, Peking University, Beijing, China\\
    \textsuperscript{\rm 2}School of Intelligence Science and Technology, Peking University, Beijing, China\\
    \textsuperscript{\rm 3}State Key Laboratory of General Artificial Intelligence, Peking University, Beijing, China\\
    \textsuperscript{\rm 4}Alibaba Group, Beijing, China\\
    \textsuperscript{\rm 5}Shenzhen International Graduate School, Tsinghua University
}

\begin{document}

\maketitle

\begin{abstract}
Diffusion models have enabled high-quality video generation in recent years, but the high cost of iterative sampling hinders their practical deployment. Few-step distillation alleviates this cost, yet exposes a quality--diversity trade-off between its two dominant paradigms: trajectory-level distillation (e.g., sCM) favors diversity, whereas distribution-level distillation (e.g., DMD) favors quality. Targeting extreme two-step video generation, we introduce \duetgradient, which reconciles the two paradigms through a noise-level duet of experts: an \scmexpert{} takes the high-noise step to lay out diverse structure, and a \dmdexpert{} takes the low-noise step to refine appearance detail. Since the two experts are trained independently with their native objectives, \method{} sidesteps the optimization difficulties of loss-level combinations and delivers quality and diversity jointly rather than trading one for the other. We further identify the relay interface and the high-noise stage as the remaining bottlenecks, and address them with RL-guided expert adaptation, yielding \duetplusgradient. With the Wan2.1-T2V-1.3B backbone, \method\ lifts the two-step quality of sCM close to the level of DMD while retaining nearly all of its structural diversity---about twice that of DMD---and \methodplus{} further improves overall quality while preserving this diversity advantage. Together, these results establish noise-level expert specialization as a simple, effective paradigm for reconciling diversity and quality in two-step video generation.
\end{abstract}

\section{Introduction}
\label{sec:introduction}

Since Sora~\citep{videoworldsimulators2024} demonstrated the potential of DiT-based~\citep{peebles2023scalable} video generation, video diffusion models have advanced at a remarkable pace: a wave of successors~\citep{wan2025wan,team2025kling,hacohen2024ltx} now generate impressively coherent clips and are seeing rapid adoption in content creation, advertising, and film production.

\begin{figure}[t]
\centering
\setlength{\fboxsep}{0.7pt}
\begin{minipage}[c]{0.15\columnwidth}
\mbox{}
\end{minipage}\hfill
\begin{minipage}[c]{0.84\columnwidth}
\centering{\scriptsize\textit{``A cute happy Corgi playing in a park, sunset, Van Gogh style.''}}
\end{minipage}\\[3pt]
\begin{minipage}[c]{0.15\columnwidth}
\centering\scriptsize
\textbf{DMD}\\[1.5pt]
{\scriptsize\colorbox{red!25}{\strut\textit{div.}}\,\colorbox{green!30}{\strut\textit{qual.}}}
\end{minipage}\hfill
\begin{minipage}[c]{0.84\columnwidth}
\includegraphics[width=\linewidth]{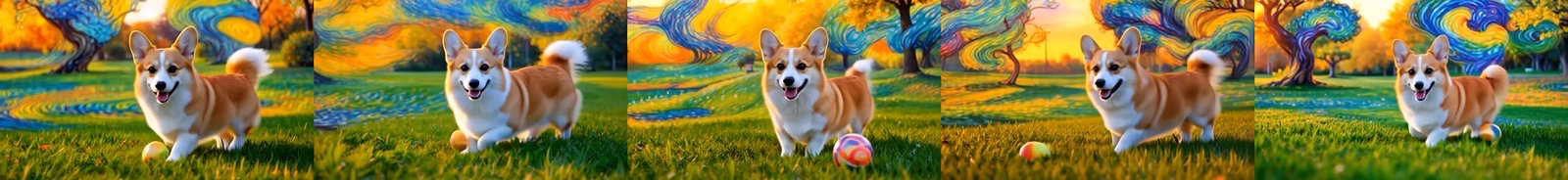}
\end{minipage}

\vspace{2pt}
\begin{minipage}[c]{0.15\columnwidth}
\centering\scriptsize
\textbf{sCM}\\[1.5pt]
{\scriptsize\colorbox{green!30}{\strut\textit{div.}}\,\colorbox{red!25}{\strut\textit{qual.}}}
\end{minipage}\hfill
\begin{minipage}[c]{0.84\columnwidth}
\includegraphics[width=\linewidth]{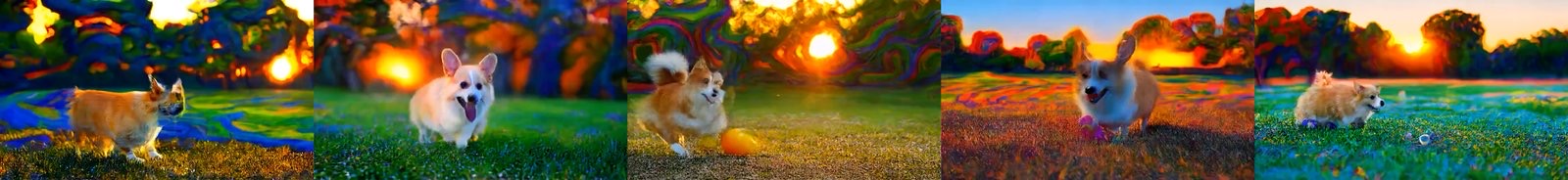}
\end{minipage}

\vspace{2pt}
\begin{minipage}[c]{0.15\columnwidth}
\centering\scriptsize
\textbf{\duetgradient}\\[1.5pt]
{\scriptsize\colorbox{green!30}{\strut\textit{div.}}\,\colorbox{green!30}{\strut\textit{qual.}}}
\end{minipage}\hfill
\begin{minipage}[c]{0.84\columnwidth}
\includegraphics[width=\linewidth]{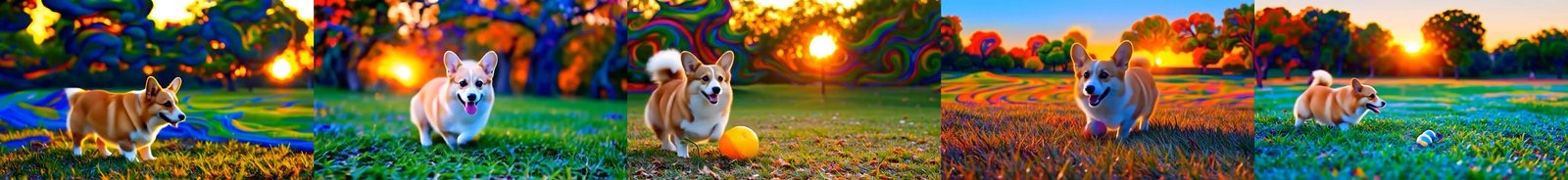}
\end{minipage}
\caption{The quality--diversity trade-off between the two distillation paradigms. Each row shows the first frames of videos generated with 5 seeds. \duetgradient{} preserves sCM's diverse layout while inheriting the high quality of DMD.}
\label{fig:qualitative-teaser}
\vspace{-3mm}
\end{figure}

However, their iterative samplers rely on long trajectories of network evaluations---producing even a single 5-second clip can take tens of minutes on a high-end GPU. Few-step distillation, which has converged on two fundamental paradigms---trajectory-level and distribution-level---addresses this cost by compressing the trajectory into few NFEs, but with no free lunch, as illustrated in Figure~\ref{fig:qualitative-teaser}. On the one hand, trajectory-level objectives, exemplified by consistency distillation~\citep{song2023consistency, lu2025simplifying}, largely preserve the teacher's sample diversity, yet their mean-seeking nature yields conservative, blurry outputs. On the other hand, distribution-level objectives, exemplified by distribution matching distillation (DMD)~\citep{yin2024one, yin2024improved}, produce sharp outputs, but their mode-seeking nature concentrates probability mass on substantially fewer modes. The quality--diversity trade-off is not merely an empirical inconvenience of few-step generation, but a symptom of the divergent natures of the two distillation objectives.

A common response to this apparent dilemma is to combine the two distillation losses into a single student training, hoping to inherit diversity from trajectory-level distillation and fidelity from distribution-level distillation~\citep{zheng2025large, cai2026mode, zou2026hiar, ge2026salt}. Yet \textit{such loss-level combinations typically find a compromise rather than a true escape}---improving one side of the trade-off still comes at the expense of the other---and introduce \textit{hard optimization} due to conflicts between the objectives. For example, rCM~\citep{zheng2025large} augments the sCM objective with a DMD loss down-weighted to 1/100, but its diversity remains substantially lower than that of its sCM counterpart, as shown in Table~\ref{tab:main}. Similarly, HiAR~\citep{zou2026hiar} requires a dedicated design to mitigate conflicts between its trajectory-matching and distribution-matching losses. These methods ask one set of parameters to negotiate between objectives with different preferences; the negotiation can be improved, but it is difficult to remove.

In this paper, we pursue a route that maintains \textit{simplicity}, yet \textit{preserves the advantages of both paradigms as much as possible}. Motivated by expert specialization in Wan2.2's MoE design and by the distinct roles of different distillation objectives, we propose \duetgradient{} (\textbf{D}iversity--q\textbf{U}ality \textbf{E}xpert \textbf{T}andem), built on a principle of \emph{noise-level expert duet}: each interval of the sampling schedule is assigned to the objective whose optimum best matches the role of that interval. Concretely, \method{} routes the high-noise interval $[1,\tau]$ to a diversity-preserving \scmexpert{}, where layout, composition, and motion are primarily determined, and the low-noise interval $[\tau,0]$ to a fidelity-oriented \dmdexpert{}, where visual details and appearance are refined. The principle is not tied to a specific step budget; we stress-test it in the extreme two-step regime, where routing is cleanest---each regime is exactly one step---and the trade-off bites hardest. As shown in Figure~\ref{fig:qualitative-teaser}, this simple expert duet preserves much of sCM's layout diversity while maintaining the visual quality of DMD.

Diverse outputs make downstream preference optimization effective. We therefore further introduce a lightweight RL-based expert adaptation for \method{}, yielding \duetplusgradient{}. Rather than applying a generic objective to the whole \method{}, we optimize each expert according to its own role. For the \scmexpert{}, whose diverse outputs provide the exploration needed for preference learning, we apply GRPO~\citep{shao2024deepseekmath, lu2026raven} to steer its high-noise predictions toward higher-reward structures while preserving its coverage behavior. For the \dmdexpert{}, we continue DMD training on the \scmexpert{}'s latents, so that it adapts to the actual intermediate distribution produced by the \scmexpert{} at inference. Experiments show that this role-aware adaptation is much more effective than the common init-then-DMD recipe~\citep{yin2025slow}, which quickly collapses the inherited diversity; \methodplus{} substantially improves image quality while preserving the diversity advantage.

To summarize, our contributions are fourfold:
\begin{itemize}
    \item We diagnose the quality--diversity trade-off at the objective level: trajectory- and distribution-level objectives induce distinct noise-to-data mappings, making loss-level combination optimization-hard and compromise-prone.
    \item We propose \duetgradient{}, a noise-level expert duet that assigns diversity- and fidelity-oriented experts to the noise regimes where they are most effective.
    \item We introduce a role-aware adaptation that exploits the \scmexpert{}'s diversity for preference optimization and repairs the \dmdexpert{}'s relay interface, yielding \duetplusgradient{}.
    \item On Wan2.1-1.3B, we show that quality and diversity can be obtained jointly in the extreme two-step regime: \method{} approaches DMD-level quality while preserving much of sCM's diversity, and \methodplus{} reaches DMD-level overall quality while preserving the diversity advantage.
\end{itemize}
\section{Related Work}
\label{sec:related}

\paragraph{Trajectory distillation.}
Trajectory-level distillation accelerates sampling by learning shortcut mappings along the teacher's ODE trajectories. Early methods such as progressive distillation~\citep{salimans2022progressive} progressively distill the DDIM~\citep{song2020denoising} sampling process, enabling shorter generation trajectories. Consistency models~\citep{song2023consistency} learn a consistency function that maps different points on the same ODE trajectory to a shared clean-data endpoint. Subsequent works extend this idea to continuous-time formulations sCM~\citep{lu2025simplifying}, multi-step, phased, and truncated sampling schemes~\citep{heek2024multistep, ren2024hyper, wang2024phased, lee2024truncated}. Another line of work directly learns flow maps between arbitrary endpoint pairs~\citep{kim2024consistency, geng2026mean, sabour2026align, frans2025one, chen2025sana, zheng2025large}. Despite these differences, these methods share the goal of matching the teacher's trajectory structure.

\paragraph{Distribution matching distillation.}
Another family of methods does not attempt to reproduce individual samples along the teacher's trajectory exactly, but instead aims to match the teacher at the distribution level. Among them, DMD~\citep{yin2024one,yin2024improved} is a representative approach, which minimizes an approximate reverse KL divergence between the student and teacher distributions with the aid of an auxiliary fake-score network. Follow-up works replace the reverse KL with alternative distribution distances, including Fisher-divergence-style objectives built on score identities~\citep{zhou2024score,luo2024one} and general $f$-divergences~\citep{xu2025one}. Beyond explicit distribution distances, adversarial distillation instead aligns the student with the teacher or data distribution through a learned discriminator~\citep{sauer2024adversarial,sauer2024fast,lin2025diffusion,lin2026autoregressive}. Distribution matching has been adopted well beyond text-to-image synthesis, spanning autoregressive and interactive video generation~\citep{yin2025slow,huang2026self,zhu2026causal}, molecule generation~\citep{wei2026flashmol}, and reward-aware distillation~\citep{jiang2025distribution,bai2026optimizing}.

\paragraph{Balancing diversity and quality.}
The mode-seeking behavior of DMD and the mode-covering tendency of trajectory distillation motivate researchers to seek a balance between quality and diversity. One line of work combines the two types of objectives, often viewed as reverse-KL- and forward-KL-style supervision. Some methods use sCM as the dominant objective and add a small amount of DMD regularization~\citep{zheng2025large}, while others treat DMD as the primary objective and introduce forward-KL-style supervision~\citep{zou2026hiar, cai2026mode, ge2026salt}. These approaches either require careful tuning of loss coefficients or must mitigate strong gradient conflicts through specialized gradient-routing designs~\citep{zou2026hiar,cai2026mode}. Another line of work stays within distribution matching and mitigates diversity collapse from inside: $f$-distill~\citep{xu2025one} selects divergences with weaker mode-seeking tendencies, AMD~\citep{bai2026optimizing} detects collapsed modes with reward proxies and applies repulsive corrections from the fake-score model, and DP-DMD~\citep{wu2026diversity} reserves the first step for regression and applies DMD only afterward. In contrast, our method avoids directly mixing the two objectives within a single training loss, and preserves diversity structurally through noise-level expert duet.

We provide additional related work about noise-level expert specialization in Appendix~\ref{sec:appendix-related}.

\section{Preliminaries}
\label{sec:preliminaries}

\subsection{Flow Matching}

Let $c$ denote a text condition, $x_0\sim\pdata(\cdot\mid c)$ a clean video latent, and $x_1\sim\Normal(0,I)$ Gaussian noise; the final video is obtained by decoding a clean latent with the decoder $\mathrm{Dec}(\cdot)$. Flow matching~\citep{lipman2022flow} connects data and noise through the linear interpolation
\begin{equation}
    x_t=(1-t)x_0+t x_1, \qquad t\in[0,1],
    \label{eq:path}
\end{equation}
and trains a teacher velocity field $v_\psi(x_t,t,c)$ by regressing the path velocity, $\min_\psi\E\!\left[\|v_\psi(x_t,t,c)-(x_1-x_0)\|_2^2\right]$. The teacher defines the probability-flow ODE
\begin{equation}
    \frac{d x_t}{dt}=v_\psi(x_t,t,c).
    \label{eq:ode}
\end{equation}
Integrating Eq.~\eqref{eq:ode} backward from $t=1$ to $t=0$ transports the noise distribution to the data distribution. Throughout this paper, $\psi$ denotes the teacher, and $\theta$ and $\phi$ parameterize the two distilled students introduced below.

\subsection{Trajectory-Level Distillation}

Trajectory-level distillation learns shortcuts of the teacher ODE~\citep{salimans2022progressive, song2023consistency, frans2025one, lu2025simplifying, zheng2025large}. Among these, consistency distillation is a representative method: a consistency function $f_\theta^{\mathrm{sCM}}(x_t,t,c)$ predicts the clean endpoint of the teacher trajectory passing through $x_t$, supervised by enforcing consistent predictions at two nearby states on the same trajectory~\citep{song2023consistency}. sCM~\citep{lu2025simplifying} takes the continuous-time limit of this objective; abstracting away parameterization-specific coefficients, its training gradient reduces to
\begin{equation}
 \nabla_\theta\mathcal{L}_{\mathrm{CM}}(\theta)=
 \nabla_\theta\,\E\!\left[w(t)\,
 f_\theta^{\mathrm{sCM}}(x_t,t,c)^{\top}
 \frac{\mathrm{d} f_{\theta^-}^{\mathrm{sCM}}}{\mathrm{d} t}\right],
 \label{eq:cm}
\end{equation}
where $\theta^-$ denotes a stop-gradient copy, $w$ is a time-dependent weight, and the tangent $\frac{\mathrm{d} f_{\theta^-}^{\mathrm{sCM}}}{\mathrm{d} t}=\partial_t f_{\theta^-}^{\mathrm{sCM}}+\nabla_{x_t} f_{\theta^-}^{\mathrm{sCM}}\!\cdot v_\psi(x_t,t,c)$ is the total derivative along the teacher ODE, computed efficiently via a Jacobian--vector product. Intuitively, minimizing Eq.~\eqref{eq:cm} drives $\frac{\mathrm{d} f_{\theta^-}^{\mathrm{sCM}}}{\mathrm{d} t}\!\to\!0$, so the prediction stays constant along each teacher trajectory and hence equals its clean endpoint. Because the supervision is tied to individual teacher trajectories, Eq.~\eqref{eq:cm} constrains the noise-to-data correspondence rather than only the output marginal, which underlies the empirically observed broad sample \textit{coverage} of consistency students.

For multi-step sampling, a clean prediction is re-noised to the next scheduled level via the re-noising operator
\begin{equation}
 \mathcal{R}_{s}(\hat x_0,\epsilon)=(1-s)\hat x_0+s\epsilon,
 \qquad \epsilon\sim\Normal(0,I),
 \label{eq:renoise}
\end{equation}
after which $f_\theta^{\mathrm{sCM}}$ is applied iteratively.

\subsection{Distribution-Level Distillation}

Let $p_\phi$ be the distribution of a few-step generator $g_\phi^{\mathrm{DMD}}$ and $p_\psi$ the teacher distribution. DMD~\citep{yin2024one,yin2024improved} optimizes a reverse-KL-style objective
\begin{equation}
 \mathcal{L}_{\mathrm{DMD}}(\phi)
 =D_{\mathrm{KL}}\!\left(p_\phi(\cdot\mid c)\,\Vert\,p_\psi(\cdot\mid c)\right),
 \label{eq:dmd}
\end{equation}
whose practical gradient is estimated from the difference between the teacher score and an online fake score evaluated on student samples. Since samples are drawn from $p_\phi$, missing teacher modes receive weak direct pressure; this supplies the familiar \textit{mode-seeking} intuition.

\begin{figure}[!t]
\centering
\includegraphics[width=\columnwidth]{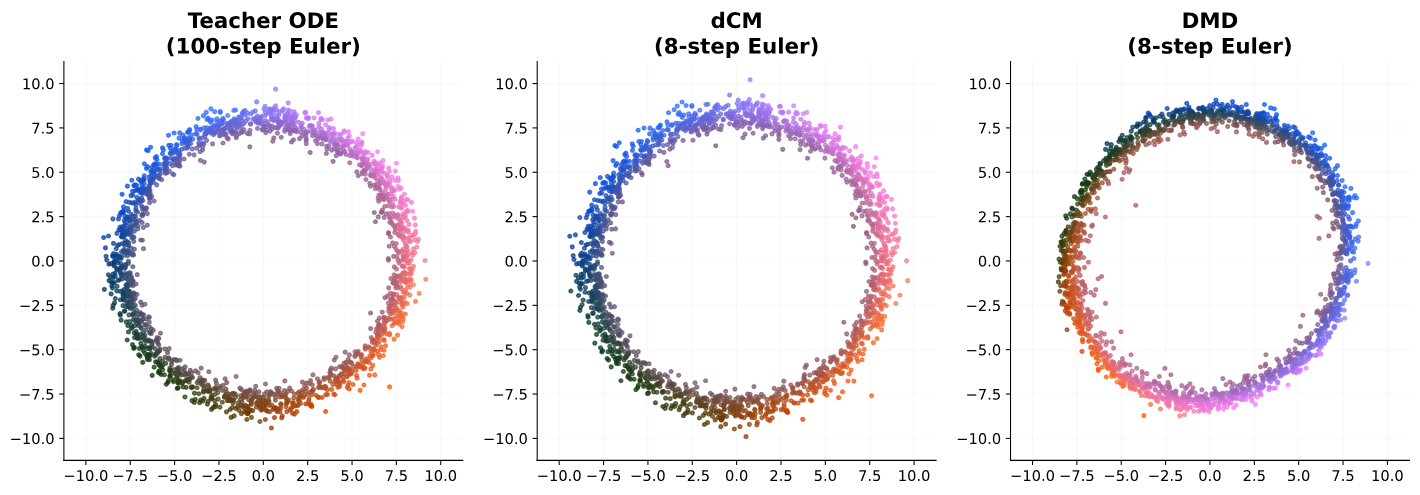}
\caption{Noise-to-data mappings of different paradigms on a toy ring-shaped Gaussian mixture with a continuum of modes. Color encodes the initial noise.}
\label{fig:mapping}
\vspace{-3mm}
\end{figure}

\begin{figure*}[!t]
\centering
\begin{minipage}[b]{0.63\textwidth}
\centering
\includegraphics[width=\linewidth]{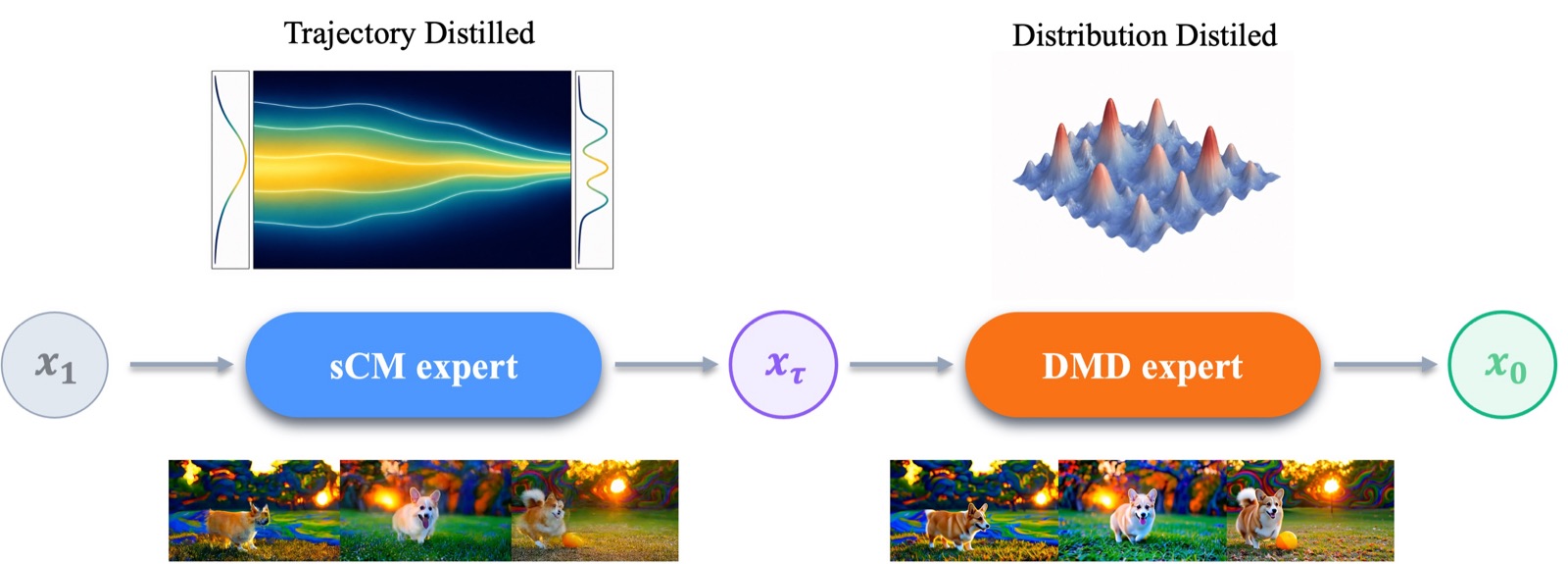}
\end{minipage}\hfill
\begin{minipage}[b]{0.33\textwidth}
\centering
\includegraphics[width=\linewidth]{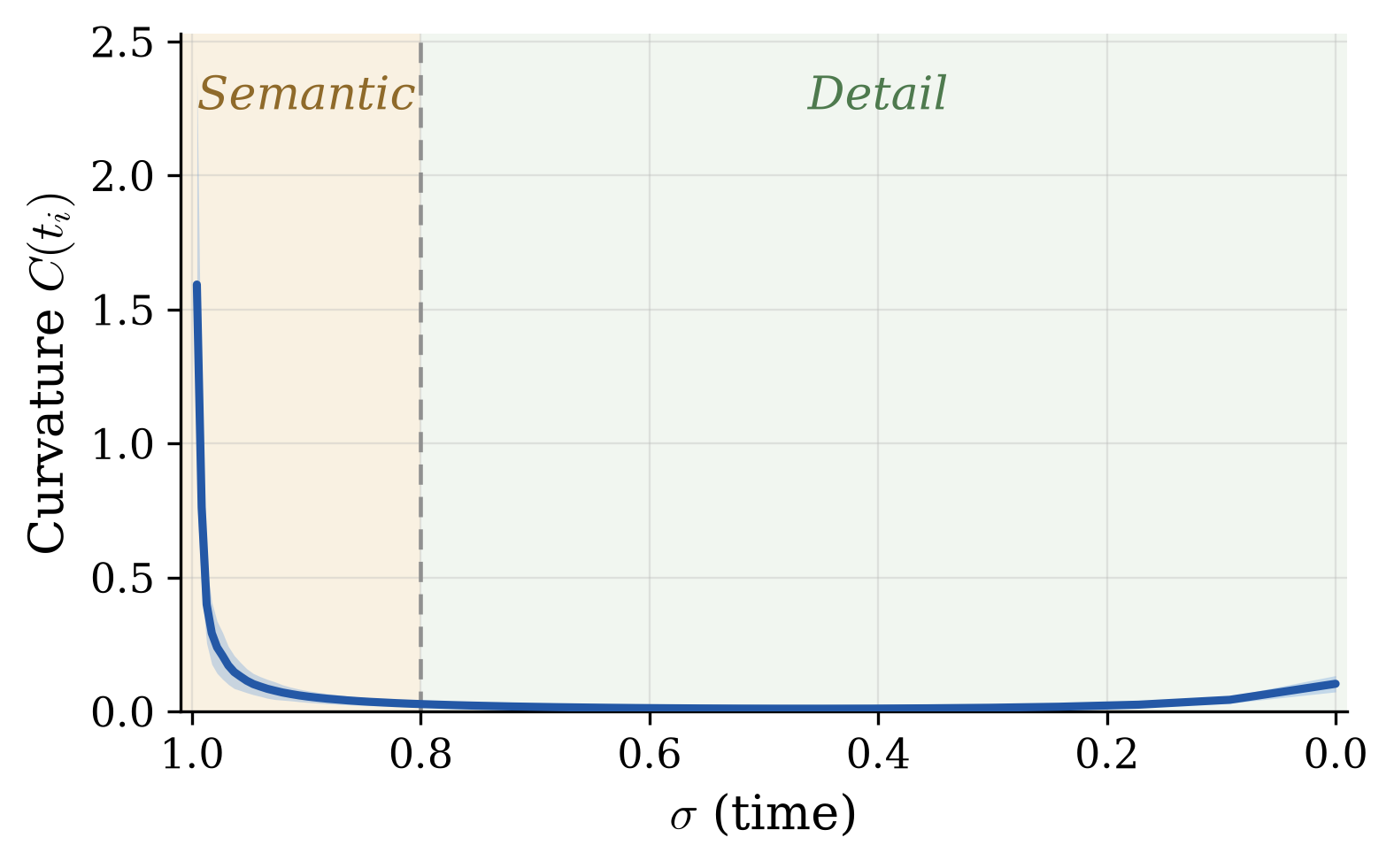}
\end{minipage}
\caption{Overview of \method{} and its switch-time selection. Left: at inference, the two independently trained experts relay within a single denoising schedule. The \scmexpert{} handles the high-noise interval $t\in[1,\tau]$ to lay out diverse structures; its prediction is then re-noised to $t=\tau$, and the \dmdexpert{} completes the low-noise interval $t\in[\tau,0]$ to refine details. Right: the curvature of the flow-matching teacher is high in the semantic-formation regime and low in the detail-refinement regime; the clear transition motivates setting $\tau=0.8$.}
\vspace{-3mm}
\label{fig:overview}
\label{fig:curvature}
\end{figure*}

\section{Method}
\label{sec:method}

\subsection{Noise-Data Mapping Conflicts}

As two fundamentally different distillation paradigms, DMD and sCM induce distinct optimization preferences: DMD seeks local shortcuts that match the output distribution, whereas sCM enforces path correspondence with the teacher's trajectory. Starting from the same teacher initialization, the two objectives therefore converge to distinct noise-to-data mappings.

We illustrate with a simple probing experiment. We first train a flow-matching teacher on a ring-shaped 2D Gaussian mixture and distill it separately with sCM and DMD. All students use deterministic 8-step Euler sampling, rather than typical consistency sampling, to preserve mapping consistency. As shown in Figure~\ref{fig:mapping}, we pass identically colored initial noise through every model and compare output coordinates, directly visualizing the noise-data correspondence. The sCM student's outputs remain closely paired with the teacher, whereas DMD realizes a very different mapping.

We hypothesize that, when the two objectives are combined, such a mismatch induces gradient conflicts that lead to hard optimization dynamics and the quality--diversity trade-off, as discussed in Section~\ref{sec:related}. This suggests that the conflict is intrinsic to forcing one set of parameters to realize both mappings, and can only be sidestepped by keeping the two objectives apart.

\subsection{\duetgradient: A Noise-Level Expert Duet}

Flow matching models naturally separate semantic formation from detail refinement across noise levels~\citep{wu2026diversity, ren2026frequency}. The high-noise regime determines coarse variables---object identity, layout, camera, and motion pattern---that shape the entire clip, while the low-noise regime has less freedom to revise this global plan and is better suited for refining edges, texture, color, and other high-frequency appearance. This division mirrors the complementary strengths of the two experts: the \scmexpert{} provides layout diversity and broad coverage, while the \dmdexpert{} improves local detail and visual fidelity.

\begin{figure}[h]
\centering
\includegraphics[width=\columnwidth]{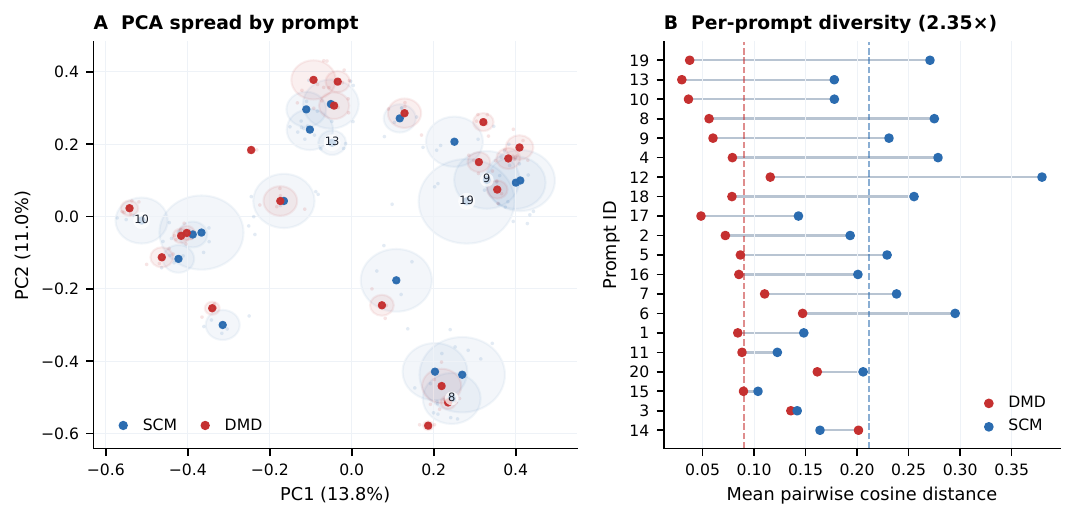}
\caption{Viclip sample embeddings of the experts.}
\label{fig:pca-spread}
\end{figure}

\paragraph{Noise-level division.}
Before starting, we first probe the two experts with a simple experiment to see the different pattern statistically: we train an \scmexpert{} and a \dmdexpert{} separately with their native losses in Eq.~\eqref{eq:cm} and Eq.~\eqref{eq:dmd}, let each generate multiple videos per prompt in a single step. As shown in Figure~\ref{fig:pca-spread}, the samples of the \scmexpert{} spread over a markedly wider region of the embedding space than those of the \dmdexpert{}---an average within-prompt spread about $2.35\times$ larger---confirming its far stronger diversity at the high-noise stage. Moreover, across distinct prompts, the two experts' sample embeddings land in nearby regions, so the distribution shift is not systematically large, laying the foundation for the relay sampling introduced next.

We therefore preserve both advantages by assigning the high-noise interval $[1,\tau]$ to a coverage-oriented \scmexpert{} $f_\theta^{\mathrm{sCM}}$ and the low-noise interval $[\tau,0]$ to a quality-oriented \dmdexpert{} $g_\phi^{\mathrm{DMD}}$, as shown in Figure~\ref{fig:overview}. In this way, quality and diversity are attained jointly rather than traded off. Since the two experts are trained independently with their native objectives, there is no coefficient to tune between $\mathcal{L}_{\mathrm{CM}}$ and $\mathcal{L}_{\mathrm{DMD}}$, and no gradient conflict to resolve.

Formally, \duetgradient{} samples in a relay manner: given text $c$ and initial noise $x_1$, the \scmexpert{} first predicts a clean endpoint $\hat x_0^{\mathrm{sCM}}=f_\theta^{\mathrm{sCM}}(x_1,1,c)$. We then re-noise this prediction to the switch time $\tau$,
\begin{equation}
    x_\tau^{\mathrm{sCM}}=\mathcal{R}_\tau(\hat x_0^{\mathrm{sCM}},\epsilon_\tau),
    \label{eq:relay-interface}
\end{equation}
and feed it to the \dmdexpert{} for the low-noise step,
\begin{equation}
 x_1\xrightarrow{\;\mathcal{R}_\tau\circ f_\theta^{\mathrm{sCM}}\;}x_\tau^{\mathrm{sCM}}
 \xrightarrow{\;g_\phi^{\mathrm{DMD}}\;}\hat x_0.
 \label{eq:relay}
\end{equation}
Algorithm~\ref{alg:relay} in Appendix~\ref{sec:appendix-algorithm} summarizes this. Note that, although two experts are involved, the sampler still uses only two network evaluations, matching the inference cost of a native two-step student while resembling the hard-routing MoE structure of Wan2.2~\citep{wan2025wan}.

\paragraph{Determining $\tau$.} The switch time $\tau$ is critical in this extremely low-budget two-step regime. A high $\tau$ (close to pure noise) leaves the semantic layout insufficiently denoised and overly blurry---a degradation the \dmdexpert{} cannot recover. Conversely, an overly low $\tau$ (close to clean data) places excessive demand on the \scmexpert{} while leaving the \dmdexpert{} little room for refinement. We choose $\tau$ based on the curvature of the flow-matching teacher's ODE trajectory~\citep{nie2026transition, feng2026one}, which is quantified as
\begin{equation}
    C(t_i)\propto\left\|
    \frac{x_{t_i}-x_{t_{i-1}}}{t_i-t_{i-1}}-(x_1-x_0)
    \right\|_2^2.
    \label{eq:curvature}
\end{equation}
Empirically, we find that in flow matching models such as Wan2.1, curvature is high near the noise endpoint, which we interpret as the fast transformation from noise to structured video, and low near the data endpoint, which we interpret as a slower refinement role. As shown in Figure~\ref{fig:curvature}, we set the switch time to $\tau=0.8$: curvature exceeds 1 before this point and drops below 0.1 afterward.

Figure~\ref{fig:qualitative-teaser} shows that, despite its simplicity, this direct relay already captures the intended division of labor: the diverse layout and motion of the \scmexpert{} are largely maintained, while the \dmdexpert{} sharpens appearance details.

\subsection{RL-Guided Expert Adaptation}
\label{sec:adaptation}

\begin{figure}[t]
\centering
\includegraphics[width=\columnwidth]{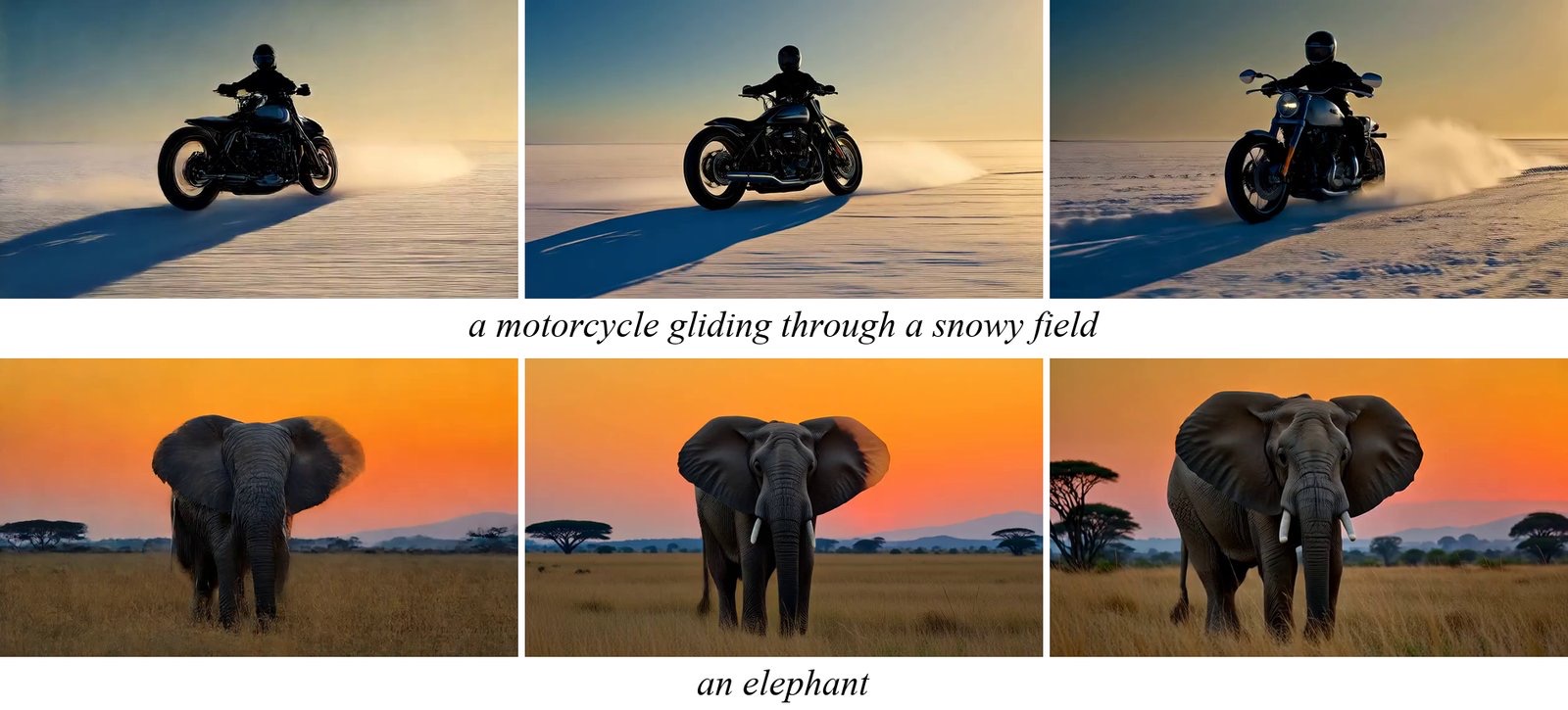}
\caption{\method{} produces high-quality videos, but the direct collaboration is still not optimal. Columns show sCM, \duetgradient{}, and \duetplusgradient{} from left to right. \methodplus{} further improves image quality and visual consistency.}
\label{fig:adaptation-necessity}
\vspace{-3mm}
\end{figure}

Although the naive expert duet already yields promising results, neither expert operates at its best, resulting in suboptimal samples, as shown in Figure~\ref{fig:adaptation-necessity}.

\begin{figure*}[!t]
\centering
\setlength{\tabcolsep}{0.5pt}
\begin{tabular}{@{}>{\centering\arraybackslash}m{0.10\textwidth}>{\centering\arraybackslash}m{0.38\textwidth}>{\centering\arraybackslash}m{0.38\textwidth}@{}}
\toprule
& \textit{``A corgi is playing drum kit''} & \textit{``Gwen Stacy reading a book, pixel art''} \\
\midrule
\textbf{DMD} &
\includegraphics[width=\linewidth]{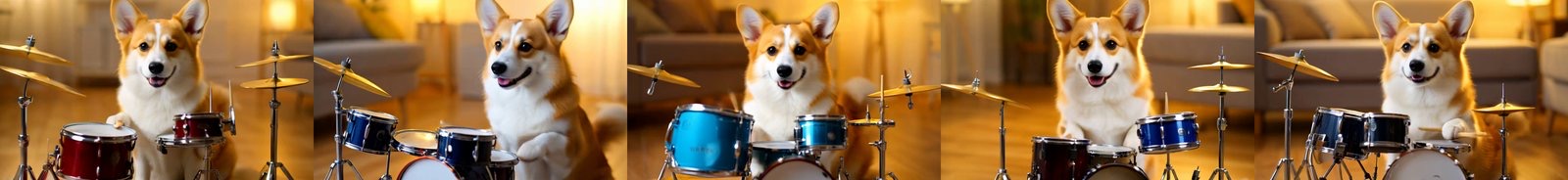} &
\includegraphics[width=\linewidth]{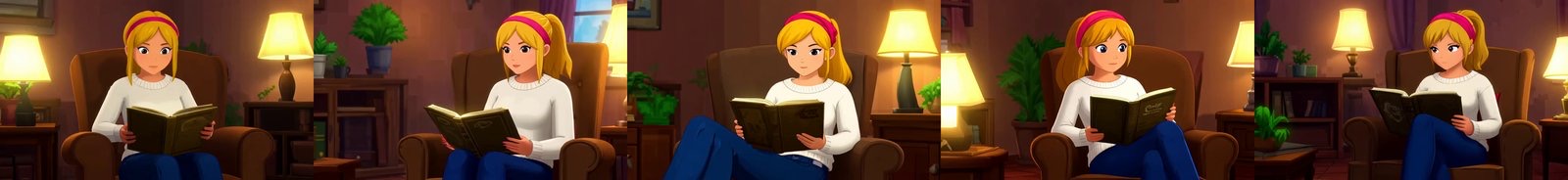} \\
\textbf{DP-DMD} &
\includegraphics[width=\linewidth]{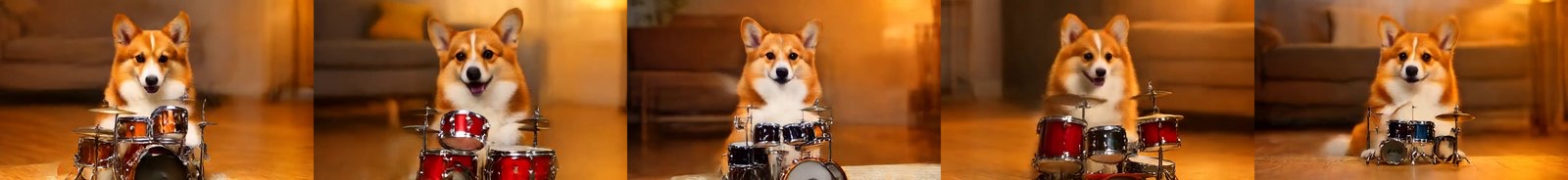} &
\includegraphics[width=\linewidth]{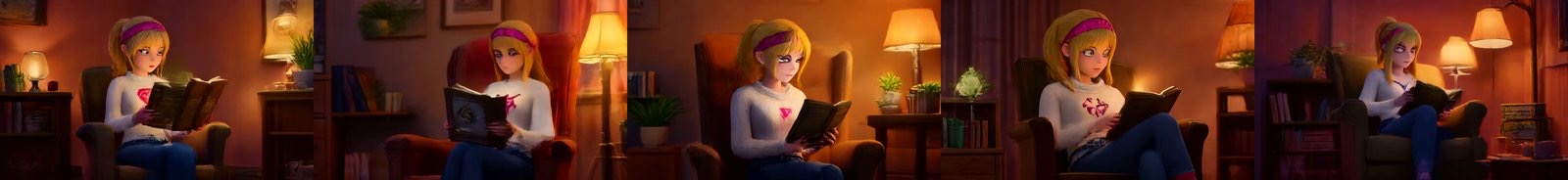} \\
\textbf{rCM} &
\includegraphics[width=\linewidth]{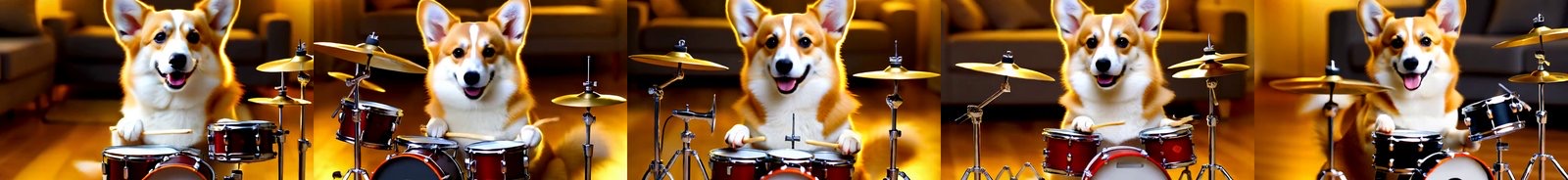} &
\includegraphics[width=\linewidth]{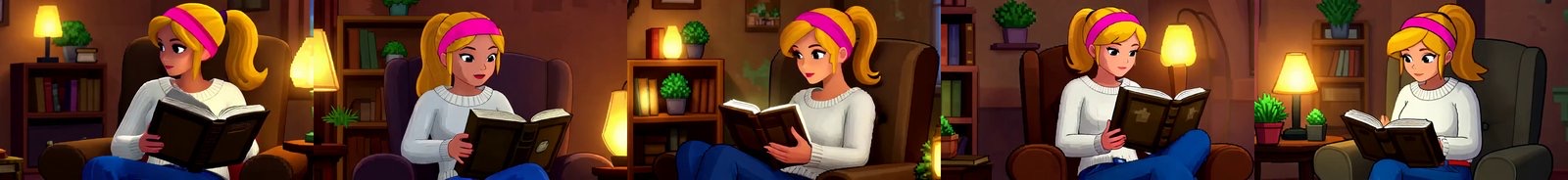} \\
\textbf{sCM} &
\includegraphics[width=\linewidth]{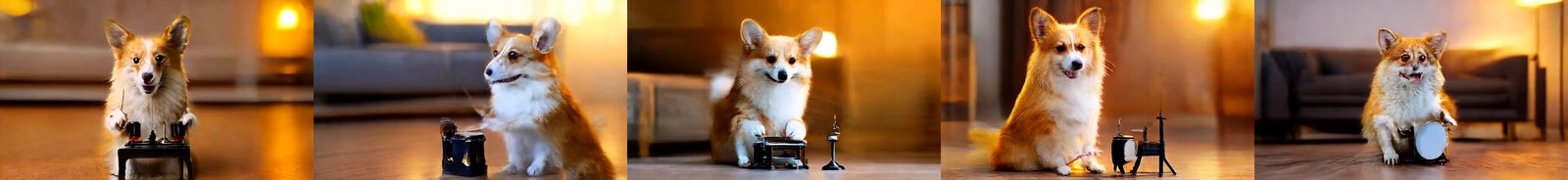} &
\includegraphics[width=\linewidth]{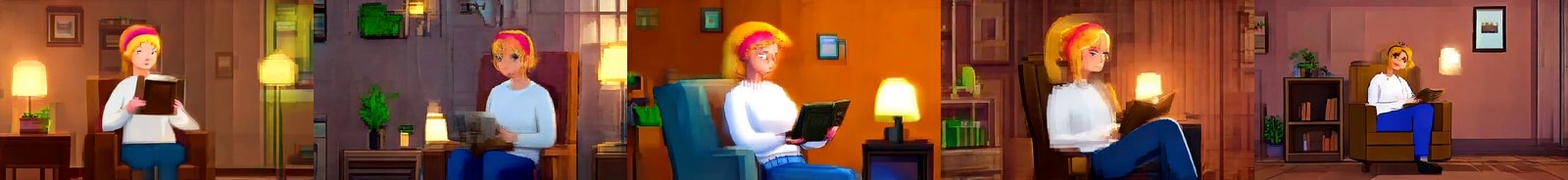} \\
\textbf{\duetgradient} &
\includegraphics[width=\linewidth]{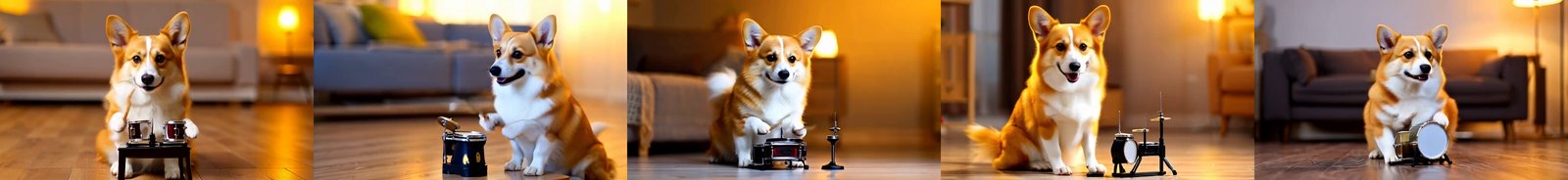} &
\includegraphics[width=\linewidth]{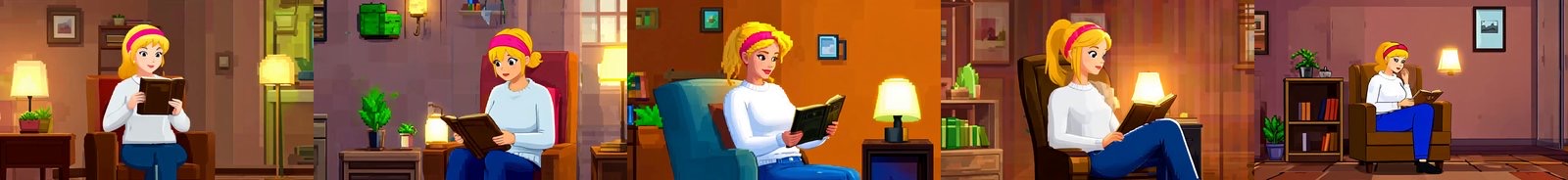} \\
{\small\textbf{\duetplusgradient}} &
\includegraphics[width=\linewidth]{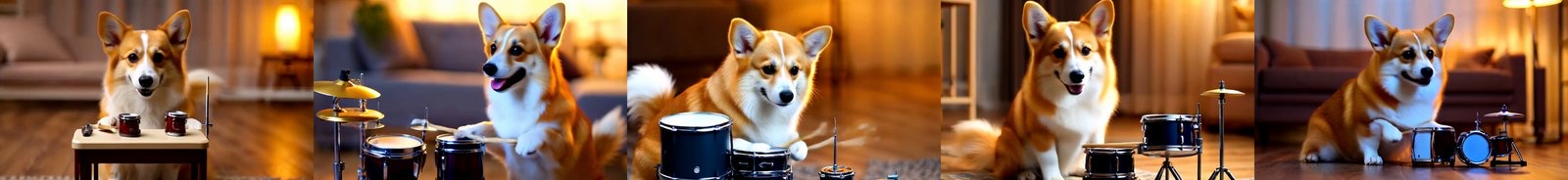} &
\includegraphics[width=\linewidth]{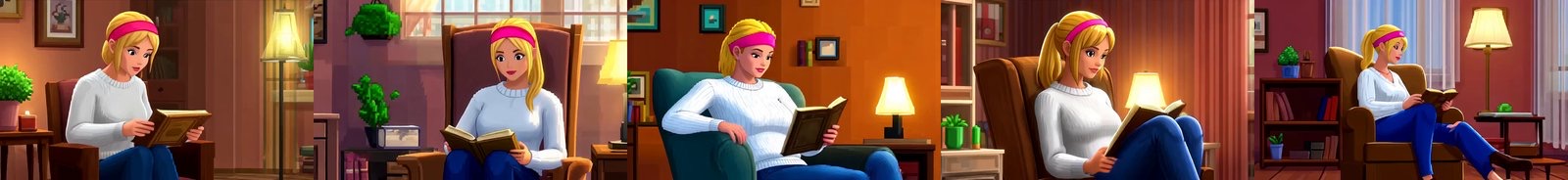} \\
\bottomrule
\end{tabular}
\caption{Qualitative comparisons across DMD, DP-DMD, rCM, sCM, \duetgradient{}, and \duetplusgradient{}. Each strip shows first frames of five generations with seeds shared across methods; within-strip variation reflects sample diversity. \textit{Zoom in for best view.}}
\label{fig:qualitative-results}
\end{figure*}

\paragraph{Remaining bottlenecks.}
We identify two limitations. First, \emph{the \dmdexpert{} faces a training--inference distribution gap}. Let $p_\tau^{\mathrm{sCM}}$ be the distribution of $x_\tau^{\mathrm{sCM}}$ produced by Eq.~\eqref{eq:relay-interface}, and let $p_\tau^{\mathrm{DMD}}$ denote that of the intermediate latents produced by the \dmdexpert{}'s native backward simulation. Ideal experts, whose predictions from any noise level follow the clean-data distribution, would give $p_\tau^{\mathrm{sCM}}\approx p_\tau^{\mathrm{DMD}}$; but finite capacity and optimization error generally yield
\begin{equation}
    p_\tau^{\mathrm{sCM}}\ne p_\tau^{\mathrm{DMD}}.
    \label{eq:mismatch}
\end{equation}
The \dmdexpert{} is thus evaluated off its training distribution. 

Second, \emph{the \scmexpert{}'s finite capacity can cap the overall performance}. In the layout-first, detail-later paradigm, we identify the layout stage as the bottleneck. This is especially true for video generation, where layout also encodes motion and therefore reflects the model's understanding of physical dynamics. The performance ceiling therefore rests on the \scmexpert{}, which is also the harder one to train: matching the teacher's trajectory endpoint requires representing the integral
$$f_{\theta}^{\mathrm{sCM},*}(x_t,t,c)=x_t+\int_t^0 v_\psi(x_s,s,c)\,\mathrm{d}s$$
of the teacher velocity field. Precisely in the high-noise interval where the \scmexpert{} operates, the teacher trajectory is highly curved (Figure~\ref{fig:curvature}), making this integral hard to fit with limited capacity. The model consequently degrades into a mean-seeking pattern, fitting an intermediate solution between distinct trajectories and sometimes producing physically implausible structures.

\paragraph{Expert-specific adaptation.}
To address these two limitations, we perform RL-guided expert adaptation for \method{}, yielding \duetplusgradient{}. We adapt each expert according to its role. 

For the \scmexpert{}, we exploit its diverse samples and use GRPO~\citep{shao2024deepseekmath, lu2026raven} to steer this distribution toward the preference distribution defined by selected rewards~\citep{ma2025hpsv3,liu2026improving}. We treat the \scmexpert{} as a one-step policy: since the relay latent is obtained by re-noising the predicted endpoint, Eq.~\eqref{eq:relay-interface} induces a Gaussian transition kernel $\pi_\theta(x_\tau^{\mathrm{sCM}}\mid x_1,c)=\Normal\big((1-\tau)\hat x_0^{\mathrm{sCM}},\tau^2 I\big)$, so policy optimization can be applied directly to the sampler used at inference without any auxiliary SDE. We therefore adopt CM-GRPO~\citep{lu2026raven}, which maximizes the advantage-weighted log-likelihood of the sampled relay transitions through the stop-gradient regression objective
\begin{equation}
\begin{aligned}
\mathcal{L}_{\mathrm{CM\text{-}GRPO}}(\theta)
&=\mathbb{E}_{c,j}\Big[\big\|\hat x_{0,j}^{\mathrm{sCM}}-\mathrm{sg}\big(\hat x_{0,j}^{\mathrm{sCM}}+\Delta_j\big)\big\|_2^2\Big],\\
\Delta_j&=\frac{(1-\tau)\hat A_j}{2\tau^2}\Big(x_{\tau,j}^{\mathrm{sCM}}-(1-\tau)\hat x_{0,j}^{\mathrm{sCM}}\Big),
\end{aligned}
\label{eq:cmgrpo}
\end{equation}
where $\hat x_{0,j}^{\mathrm{sCM}}=f_\theta^{\mathrm{sCM}}(x_1^j,1,c)$ is the endpoint prediction of the $j$-th sampled trajectory and $\mathrm{sg}(\cdot)$ denotes stop-gradient. The advantage $\hat A_j$ is the group-normalized reward. The gradient of Eq.~\eqref{eq:cmgrpo} recovers exactly the advantage-weighted score of the Gaussian relay kernel, reinforcing endpoint predictions whose relay latents lead to high-reward videos. 

For the \dmdexpert{}, we continue training with the DMD loss, but construct the backward simulation with \method{} instead of the native rollout. Notably, adaptation proceeds within a single training run: we interleave updates of the \scmexpert{} and \dmdexpert{}, so that the \dmdexpert{} always adapts to the latest output distribution of the \scmexpert{}.

Direct comparison in Figure~\ref{fig:adaptation-necessity} demonstrates the effectiveness of this adaptation, with comprehensive quantitative results provided in Section~\ref{sec:experiments}.
\section{Experiments}
\label{sec:experiments}
\begin{table*}[t]
\centering
\small
\setlength{\tabcolsep}{2pt}
\begin{tabular}{@{}l|ccc|c|ccccccc|c@{}}
\toprule
& \multicolumn{4}{c|}{Diversity $\uparrow$} & \multicolumn{8}{c}{Quality $\uparrow$} \\
\cmidrule(lr){2-5} \cmidrule(l){6-13}
Method & ViCLIP & DINO & CLIP & Average & SC & BC & TF & MS & DD & AQ & IQ & Average \\
\midrule
sCM & \first{.2085} & \first{.2554} & \first{.0955} & \first{.1865} & 89.26 & 91.75 & \third{98.19} & 95.37 & 55.83 & 60.69 & 66.63 & 81.51 \\
DMD & .0767 & .1002 & .0412 & .0727 & \first{94.95} & \second{93.30} & 96.90 & \second{96.05} & 65.79 & \first{66.05} & \first{68.35} & \second{84.38} \\
\midrule
rCM & .0793 & .1105 & .0452 & .0783\,{\textcolor{red!70!black}{$\scriptstyle(+7.7\%)$}} & \third{93.55} & 92.86 & 97.36 & 93.52 & \first{73.89} & \third{65.53} & \third{67.96} & \third{84.26}\,{\textcolor{red!70!black}{$\scriptstyle(+2.75)$}} \\
DP-DMD & .1183 & .1472 & .0574 & .1076\,{\textcolor{red!70!black}{$\scriptstyle(+48.0\%)$}} & 92.04 & 92.13 & 97.95 & 95.20 & 48.33 & 62.12 & 62.47 & 80.93\,{\textcolor{green!50!black}{$\scriptstyle(-0.58)$}} \\
\midrule
\duetgradient & \third{.1600} & \second{.2102} & \third{.0834} & \third{.1512}\,{\textcolor{red!70!black}{$\scriptstyle(+108.0\%)$}} & 91.90 & \first{93.41} & \first{98.38} & \third{95.51} & \second{67.50} & 64.94 & 67.85 & 83.96\,{\textcolor{red!70!black}{$\scriptstyle(+2.45)$}} \\
\duetplusgradient & \second{.1620} & \third{.2081} & \second{.0861} & \second{.1521}\,{\textcolor{red!70!black}{$\scriptstyle(+109.2\%)$}} & \second{94.13} & \third{93.23} & \second{98.35} & \first{96.19} & \third{65.83} & \second{65.61} & \second{68.14} & \first{84.40}\,{\textcolor{red!70!black}{$\scriptstyle(+2.89)$}} \\
\bottomrule
\end{tabular}
\caption{Quantitative Results. All methods use exactly two sampling steps (NFE\,$=$\,2). Parentheses report relative diversity gains over \textit{DMD} and quality point gains over \textit{sCM}. Cells shaded in \colorbox{red!35}{red}, \colorbox{blue!25}{blue}, and \colorbox{green!30}{green} mark the best, second- and third-best results in each column.}
\label{tab:main}
\vspace{-3mm}
\end{table*}

In this section, we evaluate \duetgradient{} through five questions: \textbf{Q1:} Does \method{} jointly deliver DMD quality and sCM diversity? \textbf{Q2:} Does RL-guided expert adaptation bring further gains? \textbf{Q3:} How does \method{} compare with alternatives including rCM~\citep{zheng2025large} and DP-DMD~\citep{wu2026diversity}? \textbf{Q4:} How do switch time $\tau$ and reward choice affect results? \textbf{Q5:} Is \methodplus{} just a better DMD initialization, or do the distinct expert roles matter?

\subsection{Experimental Setup}

\paragraph{Model and sampling.}
To demonstrate the effectiveness of \method{}, we focus on the rather challenging setting of two-step video generation.
We adopt Wan2.1-T2V-1.3B~\citep{wan2025wan} as the teacher model, which generates 81-frame text-to-video samples at $832\times480$ resolution. For sCM training, we use the continuous JVP kernel of rCM~\citep{zheng2025large} for efficiency; for DMD training, we use the distribution-matching objective and basic training setup of DMD2~\citep{yin2024improved}. Detailed configurations are provided in Appendix~\ref{sec:appendix-training}. We use the Wan2.1-14B-synthesized data provided by~\citet{zheng2025large} for all distillation.

\paragraph{Evaluation.}
Our evaluation follows the two axes of the quality--diversity trade-off studied in this paper. For \emph{quality}, we use VBench scores as the main benchmark and report the quality-relevant dimensions together with the aggregated quality score. For \emph{diversity}, we measure same-prompt diversity following DP-DMD~\citep{wu2026diversity}, in the feature spaces of ViCLIP~\citep{wang2024internvid}, DINO~\citep{caron2021emerging}, and CLIP~\citep{radford2021learning}.

\paragraph{Baselines.}
We compare the following methods, \textit{each with just two steps}: 1) \textbf{sCM} and \textbf{DMD}, which execute two native steps and isolate the two objectives; 2) \textbf{rCM}, a loss-level consistency/score-distillation combination~\citep{zheng2025large}; 3) \textbf{DP-DMD}~\citep{wu2026diversity}, which assigns ODE regression to the first step and DMD to the second; and 4) our proposed \textbf{\method{}} and \textbf{\methodplus{}}. For fairness, all results are obtained in this work using the same evaluation pipeline.

\subsection{Qualitative results}

We first present qualitative comparisons to illustrate the effectiveness of \method{} in Figure~\ref{fig:qualitative-results}. Since the outputs are videos, we show only first frames in the main paper and provide more video frames in Appendix~\ref{sec:appendix-qual}.

We highlight two representative cases. The first prompt is subject-centric, where \duetgradient{} preserves the diverse compositions and poses produced by the \scmexpert{} while substantially improving visual fidelity through the \dmdexpert{}. The second prompt tests appearance style. For such a pixel-art prompt, DMD often drifts toward smoother animation-like videos rather than blocky pixel art, suggesting that DMD-dominated distribution matching may prefer high-probability appearance modes while ignoring lower-probability styles. In contrast, \duetgradient{} retains the \scmexpert{}'s coverage of this lower-probability style while still sharpening local details. Building on the base relay, \duetplusgradient{} further enhances the aesthetic quality and repairs residual artifacts left by the direct relay, yielding cleaner and more coherent frames. In parallel, the diversity of rCM remains constrained by the underlying quality--diversity trade-off. DP-DMD loses both fine details and part of the structural diversity: its first stage performs only a simple ODE regression, and the two roles share a single set of parameters rather than explicitly separated experts.

\subsection{Quantitative results}

Table~\ref{tab:main} quantitatively confirms the quality--diversity trade-off and the benefit of our relay design, effectively answering \textbf{Q1}, \textbf{Q2}, and \textbf{Q3}. We make three observations. 

First, sCM attains the highest same-prompt diversity but falls far behind DMD on fidelity-oriented VBench dimensions, whereas DMD achieves strong quality at the cost of severe diversity collapse. Compared with rCM and DP-DMD, \duetgradient{} achieves the largest diversity gain over DMD, raising the diversity average from .0727 to .1512 (+108.0\%); meanwhile, its quality average rises from sCM's 81.51 to 83.96 (+2.45 points). Although its diversity score is slightly lower than that of sCM, the visualizations show that \duetgradient{} preserves diverse structures and styles while producing cleaner, more coherent samples, which may lead to a more compact feature-space distribution.

Moreover, \duetplusgradient{} keeps this diversity gain (+109.2\% over DMD) and further increases the quality improvement over sCM to +2.89 points. Consistent with our qualitative observations, alternatives such as rCM and DP-DMD merely locate a sweet spot along the quality--diversity trade-off, whereas our relay-based methods attain both jointly.

\subsection{Ablations}

In this section, we answer the remaining questions (\textbf{Q4} and \textbf{Q5}) by ablating the switch time and the reward choice, and by comparing against init-then-DMD.

\paragraph{Switch Time}
\begin{table}[t]
\centering
\small
\setlength{\tabcolsep}{1.8pt}
\begin{tabular}{@{}c|ccccccc|c@{}}
\toprule
$\tau$ & SC & BC & TF & MS & DD & AQ & IQ & Quality \\
\midrule
0.3 & 89.47 & 92.85 & \first{98.52} & 95.85 & 59.17 & 59.89 & 66.01 & 81.87 \\
\cellcolor{gray!15}0.8 & 91.90 & 93.41 & 98.38 & 95.51 & \first{67.50} & \first{64.94} & \first{67.85} & \first{83.96} \\
0.934 & \first{92.48} & \first{93.60} & 97.92 & \first{96.09} & 58.89 & 64.07 & 67.30 & 83.22 \\
\bottomrule
\end{tabular}
\caption{Effect of the switch time $\tau$ on \method{}. The gray cell ($\tau=0.8$) is our default setting. Red marks the best score in each column.}
\label{tab:tau}
\end{table}

We compare a small $\tau=0.3$, the default $\tau=0.8$, and a large $\tau=0.934$. For each value of $\tau$, we retrain the corresponding \dmdexpert{} with the same switch point in its backward simulation. Table~\ref{tab:tau} reports the quantitative results, and a qualitative comparison is provided in the Appendix (Figure~\ref{fig:tau-qualitative}). When $\tau$ is small, the \scmexpert{} is overburdened and the \dmdexpert{} has little room for refinement, yielding blurry results. When $\tau$ is large, the \scmexpert{} contributes too little, leading to under-formed layouts and missing details. The default $\tau=0.8$, selected from the teacher's trajectory curvature (Figure~\ref{fig:curvature}), balances the two roles and achieves the best scores on most dimensions as well as the highest aggregated quality.

\paragraph{Reward Function}
\begin{table}[t]
\centering
\small
\setlength{\tabcolsep}{1.5pt}
\begin{tabular}{@{}l|ccccccc|c@{}}
\toprule
Reward & SC & BC & TF & MS & DD & AQ$^\star$ & IQ$^\star$ & Quality \\
\midrule
\cellcolor{gray!15} HPSv3 & \first{93.96} & \first{94.10} & 96.76 & 94.37 & 61.94 & \first{66.86} & \first{71.26} & \first{84.36} \\
SFS-Nature & 91.38 & 93.15 & 97.92 & 94.00 & \first{68.61} & 64.32 & 69.92 & 83.84 \\
TA & 93.75 & 93.84 & \first{97.98} & \first{94.65} & 57.78 & 66.41 & 68.67 & 83.72 \\
\bottomrule
\end{tabular}
\caption{Reward-function ablation for CM-GRPO on the \scmexpert{}. $\star$ marks the image-quality-related dimensions we prioritize, and the gray cell (HPSv3) is our default reward. Red marks the best score per column.}
\label{tab:reward-choice}
\vspace{-5mm}
\end{table}

We further ablate which preference reward to use for the RL update. To isolate the effect of the reward choice, in this ablation we apply CM-GRPO only to the \scmexpert{} under different rewards, without any DMD-side adaptation, and sample with \method{}. Table~\ref{tab:reward-choice} compares HPSv3~\citep{ma2025hpsv3}, SFS-Nature~\citep{wu2026seeing}, and TA~\citep{liu2026improving} on the same VBench dimensions reported in Table~\ref{tab:main}. HPSv3 is stronger on consistency, aesthetic quality, and imaging quality yet SFS-Nature yields higher Dynamic Degree. Since our primary goal is to improve the fidelity of the \scmexpert{}'s generations, we favor HPSv3 for its improvement on the image-quality deficit caused by the \scmexpert{}'s finite capacity.

\paragraph{Compared to sCM-initialized DMD.}
\begin{figure}[!t]
\centering
\includegraphics[width=\columnwidth]{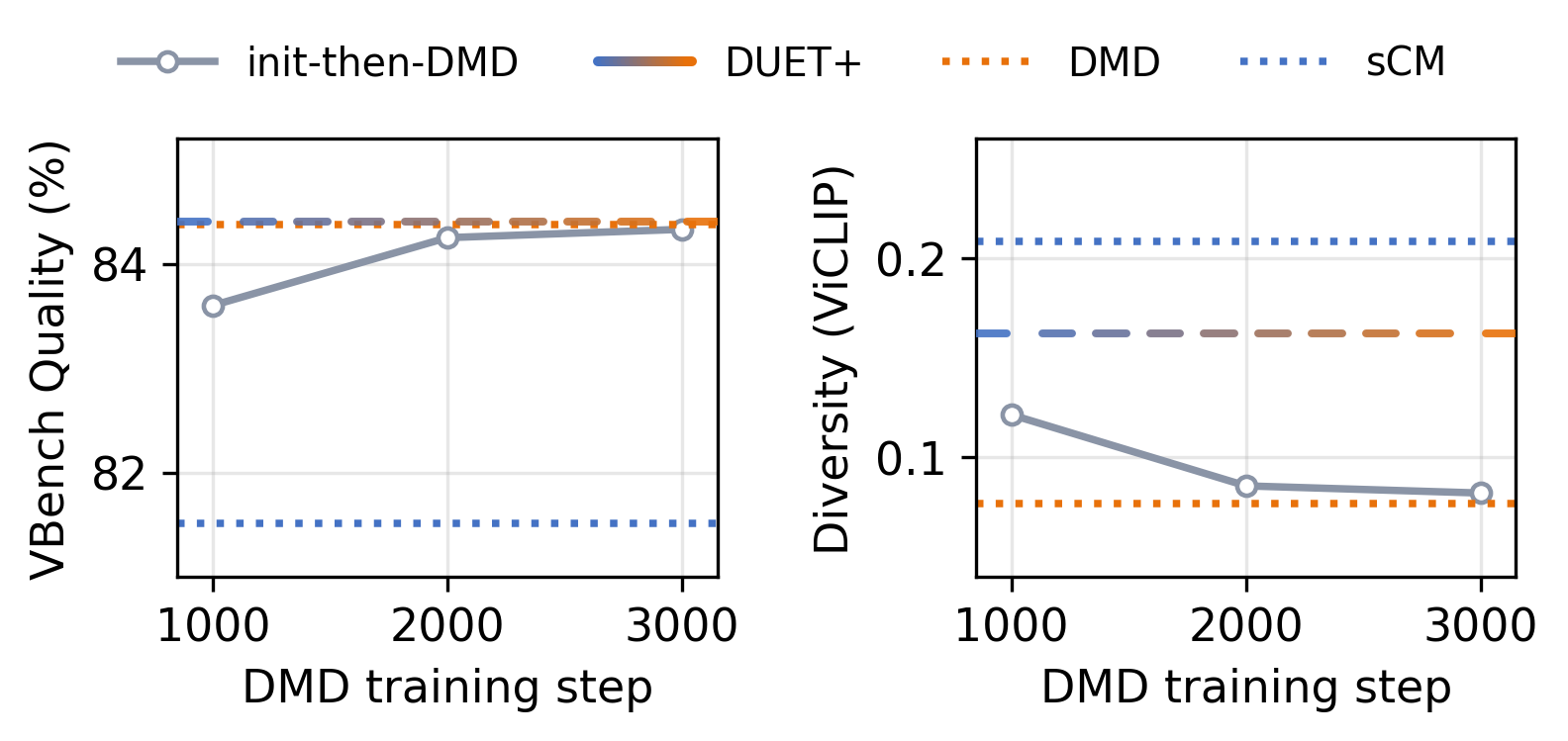}
\caption{Training dynamics of init-then-DMD, compared with \methodplus{}, DMD, and sCM. Init-then-DMD reaches comparable VBench Quality, but the diversity inherited from the sCM initialization steadily collapses toward the DMD level, remaining well below \methodplus{}.}
\label{fig:scminit-dmd}
\vspace{-3mm}
\end{figure}

\duetplusgradient{}'s adaptation resembles a common recipe in DMD-based video distillation: initialize the DMD student from a draft few-step generator---via ODE regression~\citep{yin2025slow,huang2026self} or consistency distillation~\citep{zhu2026causal}---and then continue training with the DMD objective, which we call \emph{init-then-DMD}. This mitigates DMD's local-optimum collapse. We highlight the essential difference and advantage of our method. First, \duetgradient{} is already a strong quality--diversity sampler before adaptation, rather than a warm start. Second, as shown in Figure~\ref{fig:scminit-dmd}, init-then-DMD reaches comparable VBench Quality but quickly collapses the inherited diversity---about half of \duetplusgradient{}. This reveals that the diversity advantage of \methodplus{} stems from preserving the distinct roles of the two experts, rather than a better initialization.

\section{Conclusion}
\label{sec:conclusion}

We propose \duetgradient{}, a noise-level expert duet for two-step video generation: a diversity-preserving \scmexpert{} and a fidelity-oriented \dmdexpert{} take the high- and low-noise steps, respectively. Since the two experts are trained independently with their native objectives, \method{} sidesteps both the quality--diversity trade-off and the optimization difficulty of loss-level combinations. A role-aware, RL-guided expert adaptation further yields \duetplusgradient{}, steering the \scmexpert{} toward higher-reward structures and adapting the \dmdexpert{} to the actual relay interface. On Wan2.1-T2V-1.3B, \method{} lifts the two-step quality of sCM to the level of DMD while retaining about twice DMD's diversity, and \methodplus{} reaches DMD-level quality with the diversity advantage largely intact. Ablations validate the switch time $\tau$ and the reward choice, and show that the prevalent init-then-DMD recipe instead collapses the inherited diversity. These results establish noise-level distillation expert specialization as a strong few-step paradigm for diverse and high-quality video generation. One limitation is that our evaluation is limited to Wan2.1-T2V-1.3B; however, the properties of sCM and DMD on which \method{} relies are not backbone-specific, and scaling to larger models like Wan2.1-14B is left for future work.

\section*{Acknowledgments}
We thank Kaiwen Zheng and Ruibin Li for helpful discussions.

\bibliography{references}

\onecolumn
\appendix
\section{Additional Related Work}
\label{sec:appendix-related}

\paragraph{Noise-level expert specialization.}
A series of works observe that different noise levels of the generative process demand different capabilities, and therefore specialize model capacity along the denoising schedule. eDiff-I~\citep{balaji2022ediff} trains an ensemble of expert denoisers for different noise intervals, ERNIE-ViLG~2.0~\citep{feng2023ernie} adopts a mixture of denoising experts across uniformly divided stages, and follow-up timestep-aware MoE architectures further extend this design~\citep{park2024switch, fang2024remix, cheng2025diff}; Wan2.2~\citep{wan2025wan} instantiates this idea at scale with a hard-routed high-noise/low-noise expert pair. Another line targets efficiency by assigning models of different sizes to different timesteps~\citep{pan2025t, yang2024denoising}. Nevertheless, all of these specialize capacity under a \textit{single diffusion training paradigm}, primarily to improve \textit{fidelity or efficiency} in \textit{many-step sampling}. In the distillation scenario, DCM~\citep{lv2025dual} decouples video consistency distillation into a semantic expert and a detail expert trained on high- and low-noise samples, respectively; however, it uses consistency losses for both experts and does not fully exploit the respective advantages of DMD and sCM.  \duetgradient{} is novel in that we assign \emph{two different distillation objectives} with complementary behaviors to the two noise regimes---a coverage-seeking \scmexpert{} for high-noise structure and a mode-seeking \dmdexpert{} for low-noise refinement---and relay between independently trained experts in the extreme \textit{two-step generation} regime, targeting the win-win in \textit{quality--diversity}.

\section{Method Details}
\label{sec:appendix-algorithm}

Algorithm~\ref{alg:relay} summarizes the two-step \duetgradient{} sampler and the RL-guided expert adaptation schedule described in the main paper.

\begin{algorithm}[h]
\caption{\duetgradient{} Inference and RL-Guided Expert Adaptation}
\label{alg:relay}
\begin{algorithmic}[1]
\REQUIRE text $c$, switch time $\tau$, \scmexpert{} $f_\theta^{\mathrm{sCM}}$, \dmdexpert{} $g_\phi^{\mathrm{DMD}}$, fake-score critic, sCM updates per round $N_{\mathrm{sCM}}$, critic updates per round $N_{\mathrm{critic}}$
\STATE \textbf{Two-step inference}
\STATE Sample $x_1,\epsilon_\tau\sim\Normal(0,I)$
\STATE $x_\tau^{\mathrm{sCM}}\leftarrow\mathcal{R}_\tau\big(f_\theta^{\mathrm{sCM}}(x_1,1,c),\epsilon_\tau\big)$
\STATE $\hat x_0\leftarrow g_\phi^{\mathrm{DMD}}(x_\tau^{\mathrm{sCM}},\tau,c)$
\STATE Output decoded video $\mathrm{Dec}(\hat x_0)$
\STATE \textbf{RL-guided expert adaptation}
\FOR{each adaptation round}
\STATE Update $g_\phi^{\mathrm{DMD}}$ with $\mathcal{L}_{\mathrm{DMD}}$ using relay latents from the current $f_\theta^{\mathrm{sCM}}$
\FOR{$i=1,\ldots,N_{\mathrm{sCM}}$}
\STATE Update $f_\theta^{\mathrm{sCM}}$ by CM-GRPO (Eq.~\eqref{eq:cmgrpo})
\ENDFOR
\FOR{$i=1,\ldots,N_{\mathrm{critic}}$}
\STATE Update the critic with the denoising score-matching loss on samples generated by the current relay
\ENDFOR
\ENDFOR
\RETURN adapted experts $f_\theta^{\mathrm{sCM}},g_\phi^{\mathrm{DMD}}$ as \duetplusgradient{}
\end{algorithmic}
\end{algorithm}

\section{Training and Evaluation Configurations}
\label{sec:appendix-training}

\subsection{Training Configurations}

\paragraph{Training devices.}
All training runs are conducted on 16--32 NVIDIA H800 GPUs, with gradient accumulation adjusted accordingly for the global batch size of 64. Training the \scmexpert{}, the \dmdexpert{}, and \duetplusgradient{} takes approximately 1{,}024, 360, and 128 GPU hours, respectively.

\paragraph{Common settings.}
All four methods use classifier-free guidance 5.0, timestep shift 5.0, bfloat16 precision, FusedAdamW, global batch size 64, and generator learning rate $2\times10^{-6}$. The base students, including the \scmexpert{}, \dmdexpert{}, and DP-DMD, are initialized from the teacher; \duetplusgradient{} is initialized from the two trained experts. For the methods using the DMD-side fake-score critic, we set the critic learning rate to $4\times10^{-7}$, critic weight decay to 0.01, critic updates to $N_{\mathrm{critic}}=5$, and switch time to $\tau=0.8$. Table~\ref{tab:training-config} summarizes the remaining method-specific configurations, which are further detailed as follows.

\begin{itemize}
\item \textbf{\scmexpert{} configuration.}
The \scmexpert{} is trained with the sCM objective in Eq.~\eqref{eq:cm}, LogNormal time sampling with mean 0.7 and standard deviation 1.6, and the tangent warm-up schedule of sCM~\citep{lu2025simplifying} for 1{,}000 iterations.

\item \textbf{DP-DMD configuration.}
DP-DMD follows the official recipe~\citep{wu2026diversity} with diversity weight 0.05 and $K=5$ anchor steps, keeping the same critic setting for comparability.

\item \textbf{rCM.}
The rCM baseline is not trained by us. We use the publicly released Wan2.1-T2V-1.3B 480p checkpoint from Hugging Face, linked from the official rCM repository~\citep{zheng2025large}. Note that, as stated in its model card, the released checkpoint is not trained under the same setting as the rCM paper: it is reproduced with limited synthetic data, whereas the original rCM uses high-quality internal data, and may therefore perform worse than the officially reported results. Nevertheless, the comparison remains fair, since all methods in this work are likewise trained on the same synthetic dataset.

\item \textbf{\duetplusgradient{} configuration.}
\duetplusgradient{} is initialized from the two trained experts above and optimized jointly for 500 iterations, where each cycle includes one DMD step, $N_{\mathrm{sCM}}=5$ CM-GRPO updates, and $N_{\mathrm{critic}}=5$ critic updates, following Algorithm~\ref{alg:relay}. CM-GRPO uses HPSv3 as the reward, a group size of 16 with 8 groups per policy update, and a 4-step \scmexpert{} rollout on the shifted grid, where the last transition skips to the clean endpoint and a random single transition receives the policy update; we do not use a KL loss, and advantages are clipped at 5.0. For the \scmexpert{} optimizer, we set the weight decay to 0 and Adam's $\epsilon$ to $10^{-10}$.

\end{itemize}

\begin{center}
\small
\setlength{\tabcolsep}{6pt}
\begin{tabular}{l|cccc}
\toprule
 & \scmexpert{} & \dmdexpert{} & DP-DMD & \duetplusgradient{} \\
\midrule
Training iterations & 5{,}000 & 3{,}000 & 3{,}000 & 500 \\
Adam betas & (0.9, 0.999) & (0.0, 0.999) & (0.0, 0.999) & (0.0, 0.999) \\
EMA & Yes & Yes & Yes & No \\
\bottomrule
\end{tabular}
\captionof{table}{Training configurations of the \scmexpert{}, the \dmdexpert{}, DP-DMD, and \duetplusgradient{}.}
\label{tab:training-config}
\end{center}

\subsection{Evaluation Configurations}

\paragraph{Evaluation details.}
All methods are evaluated with exactly two sampling steps (\nfes{}\,$=$\,2). For VBench, we generate videos with the prompt suites of the dimensions reported in Table~\ref{tab:main}, using the augmented prompts provided by rCM~\citep{zheng2025large}; each prompt is sampled 5 times, except for the Temporal Flickering prompts, which are sampled 25 times. For diversity, we follow DP-DMD~\citep{wu2026diversity} and compute the same-prompt diversity
\begin{equation}
    D = 1 - \frac{2}{R(R-1)}\sum_{i<j}\cos\big(x^{(i)},x^{(j)}\big),
    \label{eq:diversity}
\end{equation}
where $x^{(1)},\ldots,x^{(R)}$ are the $\ell_2$-normalized embeddings of the $R$ videos generated from the same prompt. For the DINO and CLIP feature spaces, we uniformly sample 8 frames per video, extract frame-level features, and average-pool them into a video embedding; for ViCLIP, the 8 uniformly sampled frames are directly encoded into a video-level embedding. Diversity is computed for every prompt of the full VBench prompt suite ($R=5$, or $R=25$ for the Temporal Flickering prompts) and then averaged over prompts.

\paragraph{Inference cost.}
Since the two experts of \duetgradient{} each execute exactly one of the two sampling steps, \duetgradient{} doubles the parameter count kept in memory, but its FLOPs and latency are identical to those of a native two-step student.

\paragraph{Explicit Trade-off of rCM.}
rCM exhibits an explicit quality--diversity trade-off, which can be controlled through the $\sigma_{\max}$ parameter of its EDM-style initial sampling step: a large $\sigma_{\max}$ substantially reduces diversity while improving quality, and vice versa. Figure~\ref{fig:rcm-sigmamax} visualizes the generations under the two settings: with $\sigma_{\max}=80$, the default of the official inference script, rCM trades quality for diversity, yet both its diversity and quality remain below those of \duetgradient{}. To preserve the generation quality of rCM, we report its results under $\sigma_{\max}=1600$. For our \duetgradient{}, we also take $\sigma_{\max}=1600$ for fairness.

\begin{center}
\includegraphics[width=0.95\textwidth]{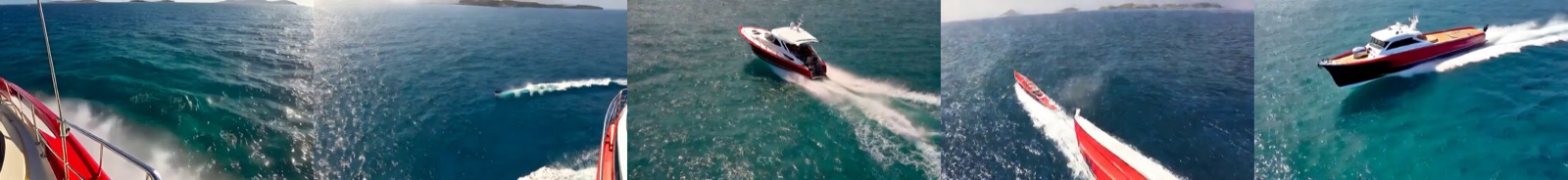}\\[2pt]
{\small rCM ($\sigma_{\max}=80$): ViCLIP diversity .1432, VBench Quality 83.22}\\[6pt]
\includegraphics[width=0.95\textwidth]{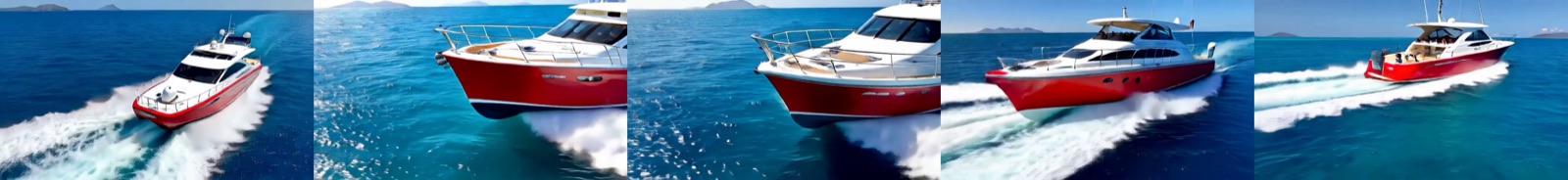}\\[2pt]
{\small rCM ($\sigma_{\max}=1600$): ViCLIP diversity .0793, VBench Quality 84.26}
\captionof{figure}{Visualization of the $\sigma_{\max}$ trade-off of rCM. Each row shows the first frames of five same-prompt generations (``A boat accelerating to gain speed'') with the corresponding scores below. $\sigma_{\max}=80$ yields diverse compositions with degraded quality, whereas $\sigma_{\max}=1600$ improves quality but collapses the generations to nearly identical layouts.}
\label{fig:rcm-sigmamax}
\end{center}

\section{More Experimental Results}
\label{sec:appendix-more}

\paragraph{Qualitative comparison across switch times.}
Figure~\ref{fig:tau-qualitative} shows samples of \duetgradient{} at different switch times $\tau$ on two prompts, using the same initial noise within each prompt block. The default $\tau=0.8$ preserves rich structural details while inheriting the high fidelity of the \dmdexpert{}; $\tau=0.934$ loses part of the fine details, and $\tau=0.3$ can hardly inherit the \dmdexpert{}'s fidelity at all.

\begin{center}
\setlength{\tabcolsep}{2pt}
\begin{tabular}{@{}>{\centering\arraybackslash}m{0.08\textwidth}>{\centering\arraybackslash}m{0.90\textwidth}@{}}
& \textit{``A panda drinking coffee in a cafe in Paris, animated style''} \\
{\small $\tau{=}0.3$} & \includegraphics[width=\linewidth]{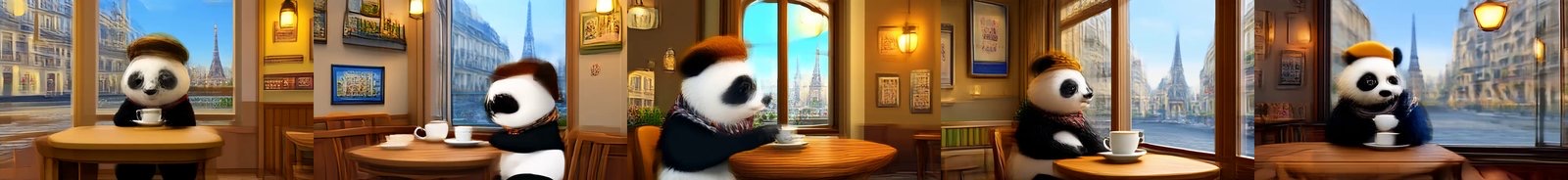} \\
{\small $\tau{=}0.8$} & \includegraphics[width=\linewidth]{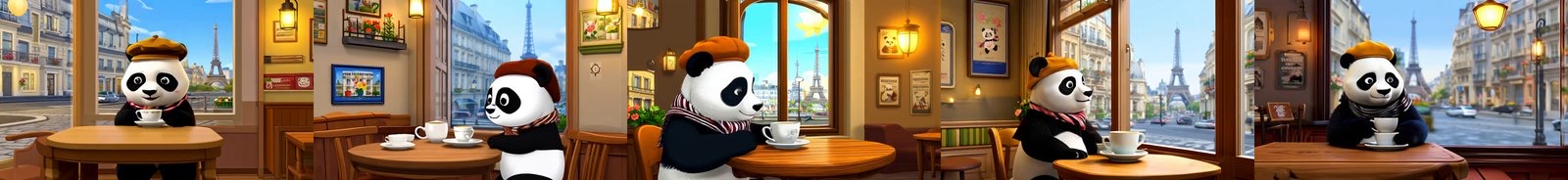} \\
{\small $\tau{=}0.934$} & \includegraphics[width=\linewidth]{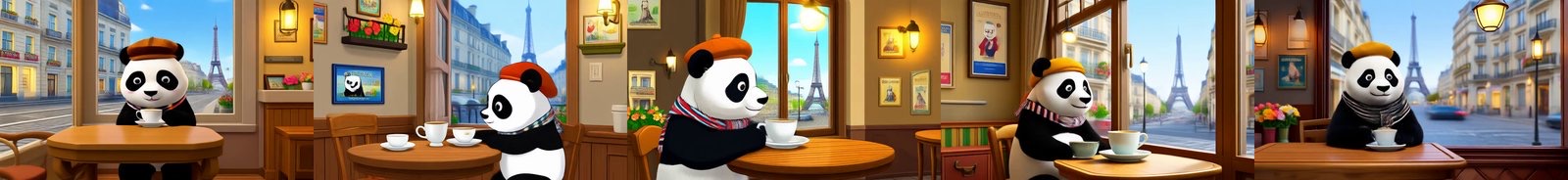} \\[4pt]
& \textit{``An astronaut flying in space, oil painting''} \\
{\small $\tau{=}0.3$} & \includegraphics[width=\linewidth]{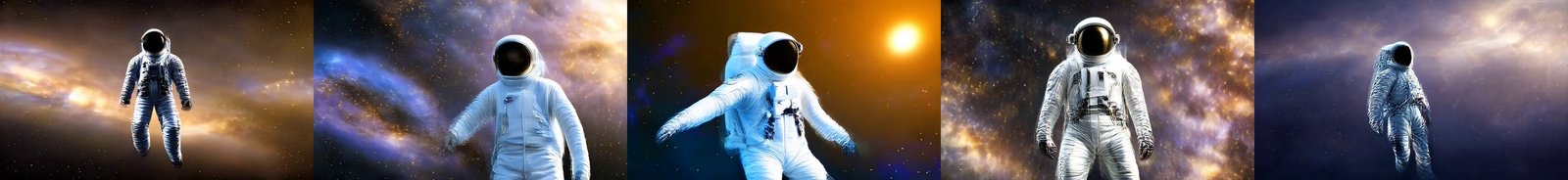} \\
{\small $\tau{=}0.8$} & \includegraphics[width=\linewidth]{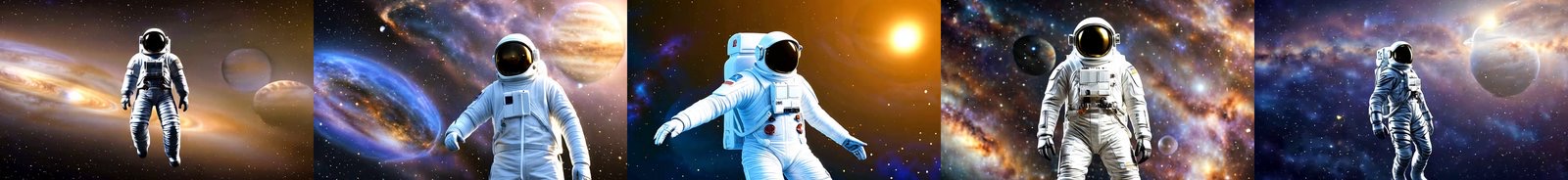} \\
{\small $\tau{=}0.934$} & \includegraphics[width=\linewidth]{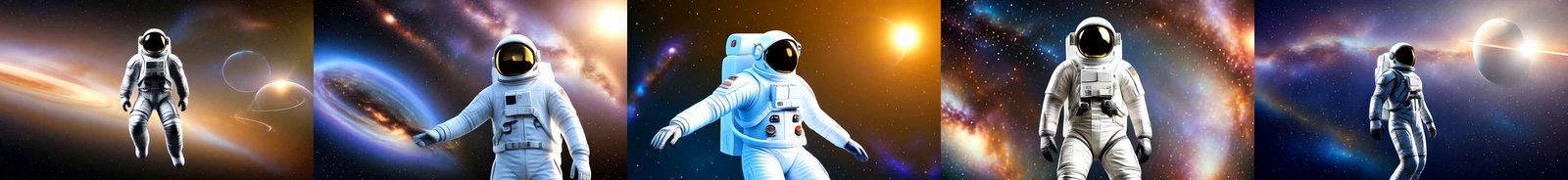} \\
\end{tabular}
\captionof{figure}{Qualitative comparison of \duetgradient{} under different switch times $\tau$ on two prompts. The default $\tau=0.8$ preserves rich structural details while inheriting the high fidelity of the \dmdexpert{}; $\tau=0.934$ loses part of the fine details, and $\tau=0.3$ can hardly inherit the \dmdexpert{}'s fidelity at all.}
\label{fig:tau-qualitative}
\end{center}

\paragraph{Reward curves of CM-GRPO.}
Figure~\ref{fig:reward-curves} plots the training reward curve of CM-GRPO applied to the \scmexpert{} with the HPSv3 reward, the setting adopted by \duetplusgradient{}. The reward mean rises steadily over the roughly 300 optimization iterations, indicating a stable preference-optimization process.

\begin{center}
\includegraphics[width=0.42\textwidth]{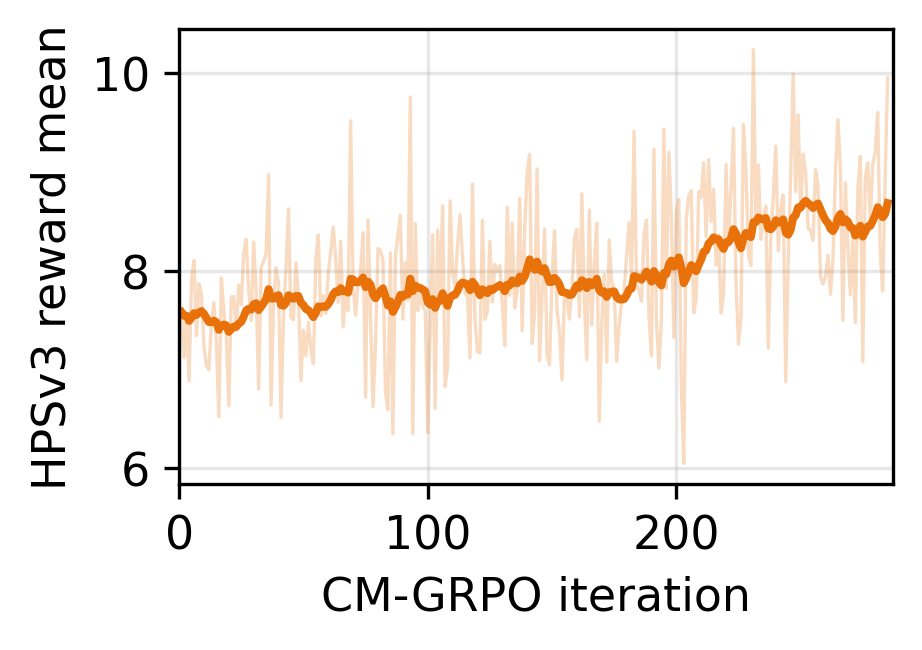}
\captionof{figure}{Training reward curve of CM-GRPO on the \scmexpert{} with the HPSv3 reward. The light curve shows the per-iteration reward mean and the dark curve shows an EMA-smoothed version.}
\label{fig:reward-curves}
\end{center}

\section{Additional Qualitative Results}
\label{sec:appendix-qual}

We provide additional qualitative results on five new prompts, covering all six models in Table~\ref{tab:main}. For each prompt, every model generates five videos with the same random seed, so the $i$-th generation of every model shares the same initial noise. Figures~\ref{fig:app-ff-corgi}--\ref{fig:app-ff-snow} show the first frames of all five generations and visualize sample diversity, while Figures~\ref{fig:app-vf-corgi}--\ref{fig:app-vf-snow} show five uniformly spaced frames of one randomly selected generation per prompt, with the generation index kept identical across models, visualizing temporal consistency.

\vspace*{\fill}

\newcommand{\appqualhdrstyled}[6]{{\normalsize\bfseries\itshape\color[HTML]{#1}\makebox[0.2\linewidth][c]{#2}\makebox[0.2\linewidth][c]{#3}\makebox[0.2\linewidth][c]{#4}\makebox[0.2\linewidth][c]{#5}\makebox[0.2\linewidth][c]{#6}}}
\newcommand{\appqualhdrseed}{\appqualhdrstyled{4472C4}{seed 0}{seed 1}{seed 2}{seed 3}{seed 4}}
\newcommand{\appqualframearrow}{\makebox[0pt][c]{$\boldsymbol{\rightarrow}$}}
\newcommand{\appqualhdrframe}{{\normalsize\bfseries\itshape\setlength{\fboxsep}{2pt}\colorbox{orange!15}{\color[HTML]{E8710A}\makebox[0.195\linewidth][c]{frame 0}\appqualframearrow\makebox[0.195\linewidth][c]{frame 20}\appqualframearrow\makebox[0.195\linewidth][c]{frame 40}\appqualframearrow\makebox[0.195\linewidth][c]{frame 60}\appqualframearrow\makebox[0.195\linewidth][c]{frame 80}}}}

{\centering
\setlength{\tabcolsep}{2pt}
\begin{tabular}{@{}>{\centering\arraybackslash}m{0.13\textwidth}>{\centering\arraybackslash}m{0.72\textwidth}@{}}
& \appqualhdrseed \\
\textbf{DMD} & \includegraphics[width=\linewidth]{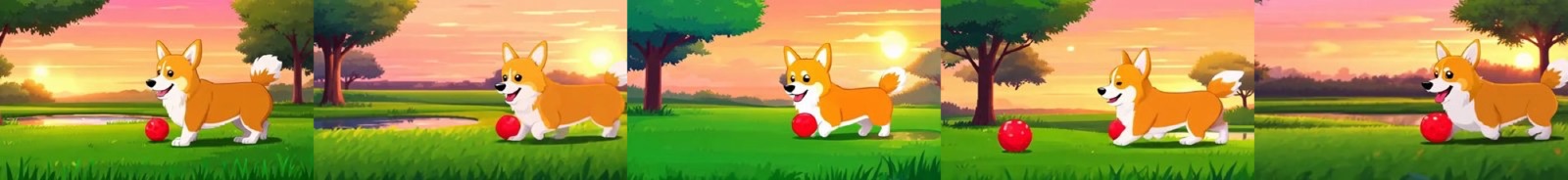} \\
\textbf{DP-DMD} & \includegraphics[width=\linewidth]{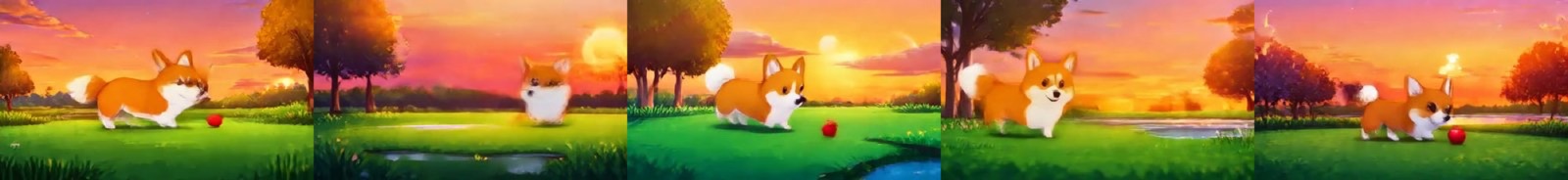} \\
\textbf{rCM} & \includegraphics[width=\linewidth]{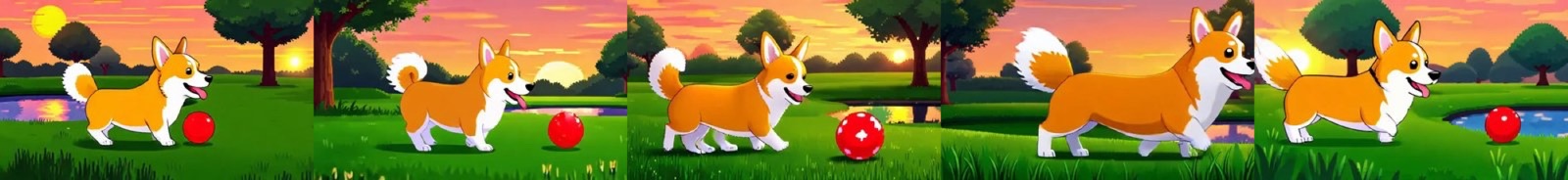} \\
\textbf{sCM} & \includegraphics[width=\linewidth]{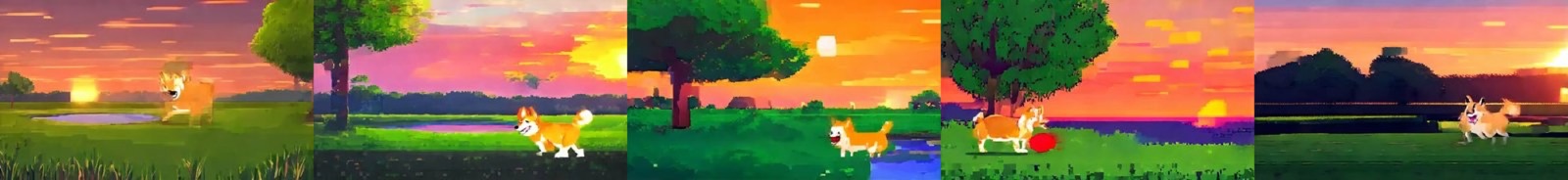} \\
\textbf{\duetgradient} & \includegraphics[width=\linewidth]{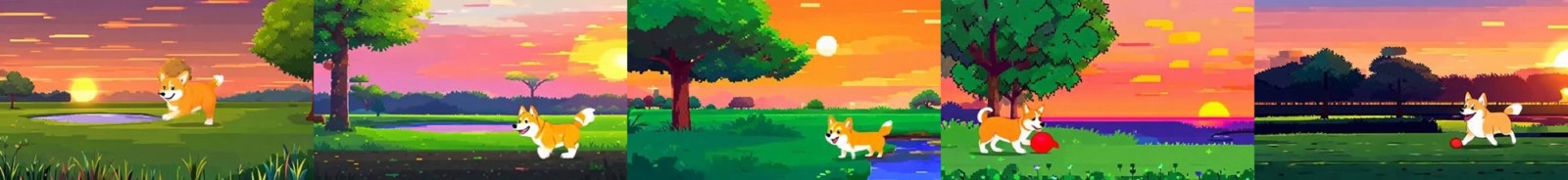} \\
\textbf{\duetplusgradient} & \includegraphics[width=\linewidth]{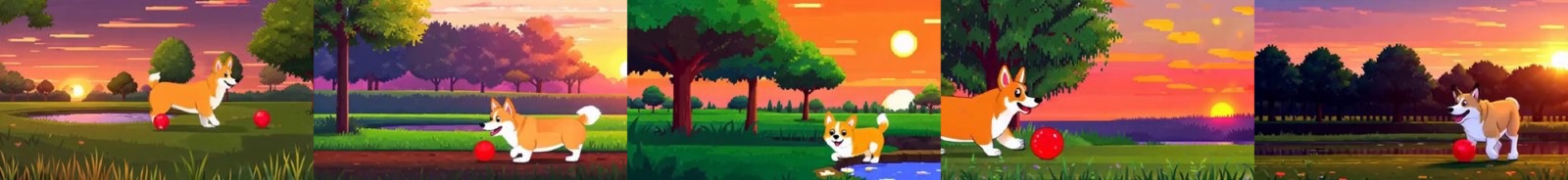} \\
\end{tabular}
\captionof{figure}{First frames of five generations on the prompt \textit{``A cute happy Corgi playing in park, sunset, pixel art''}; each column is one shared seed.}
\label{fig:app-ff-corgi}
\par}\vspace{6pt}

{\centering
\setlength{\tabcolsep}{2pt}
\begin{tabular}{@{}>{\centering\arraybackslash}m{0.13\textwidth}>{\centering\arraybackslash}m{0.72\textwidth}@{}}
& \appqualhdrseed \\
\textbf{DMD} & \includegraphics[width=\linewidth]{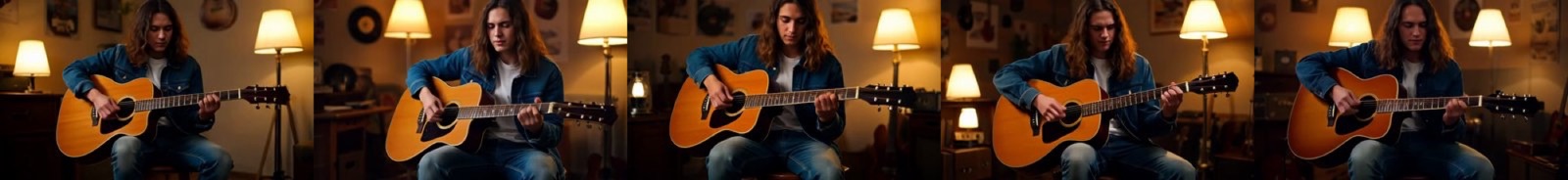} \\
\textbf{DP-DMD} & \includegraphics[width=\linewidth]{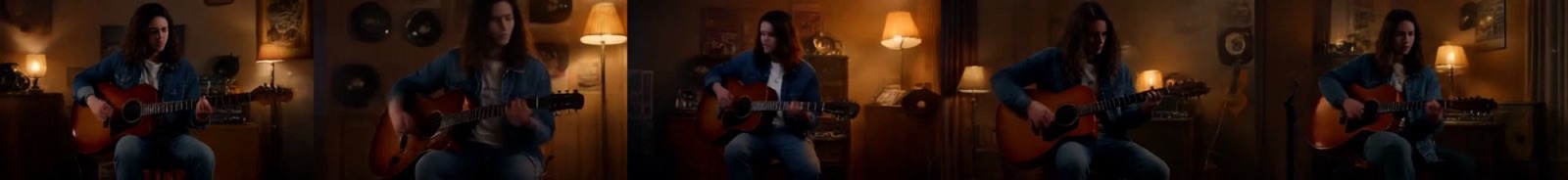} \\
\textbf{rCM} & \includegraphics[width=\linewidth]{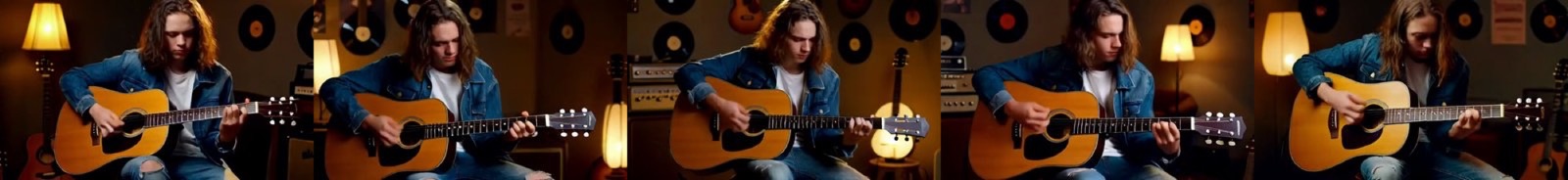} \\
\textbf{sCM} & \includegraphics[width=\linewidth]{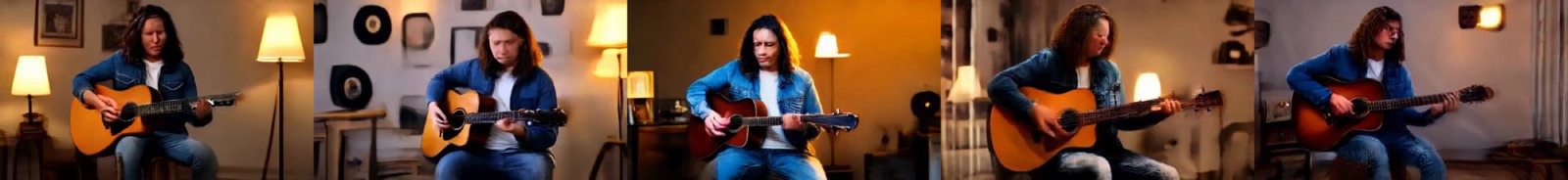} \\
\textbf{\duetgradient} & \includegraphics[width=\linewidth]{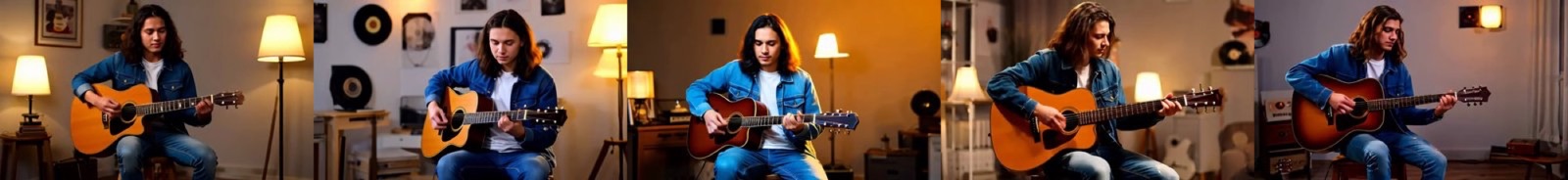} \\
\textbf{\duetplusgradient} & \includegraphics[width=\linewidth]{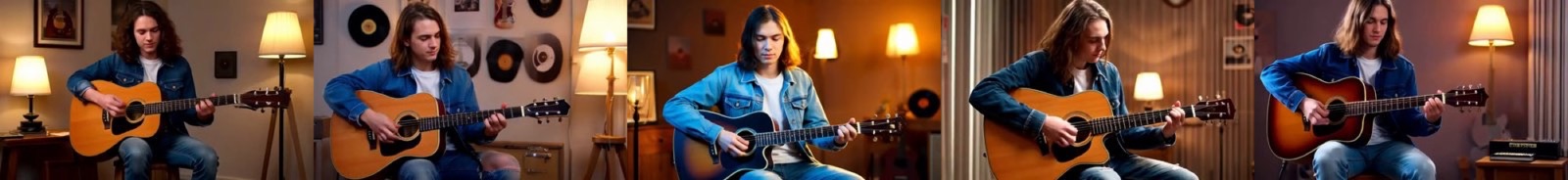} \\
\end{tabular}
\captionof{figure}{First frames of five generations on the prompt \textit{``A person playing guitar''}; each column is one shared seed.}
\label{fig:app-ff-guitar}
\par}\vspace{6pt}

{\centering
\setlength{\tabcolsep}{2pt}
\begin{tabular}{@{}>{\centering\arraybackslash}m{0.13\textwidth}>{\centering\arraybackslash}m{0.72\textwidth}@{}}
& \appqualhdrseed \\
\textbf{DMD} & \includegraphics[width=\linewidth]{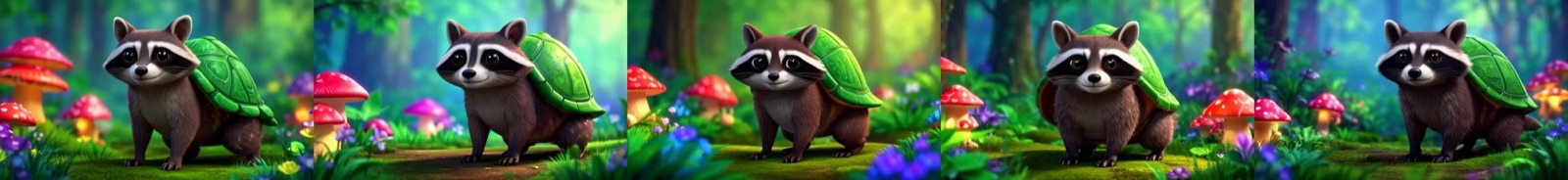} \\
\textbf{DP-DMD} & \includegraphics[width=\linewidth]{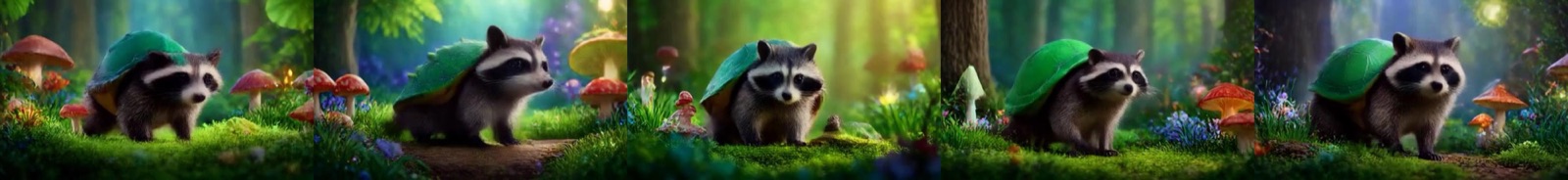} \\
\textbf{rCM} & \includegraphics[width=\linewidth]{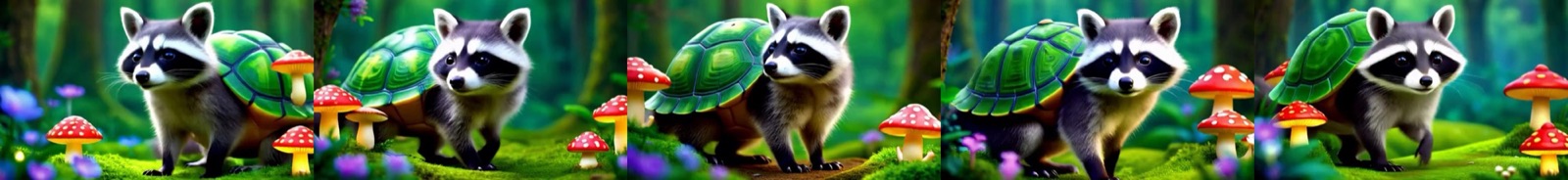} \\
\textbf{sCM} & \includegraphics[width=\linewidth]{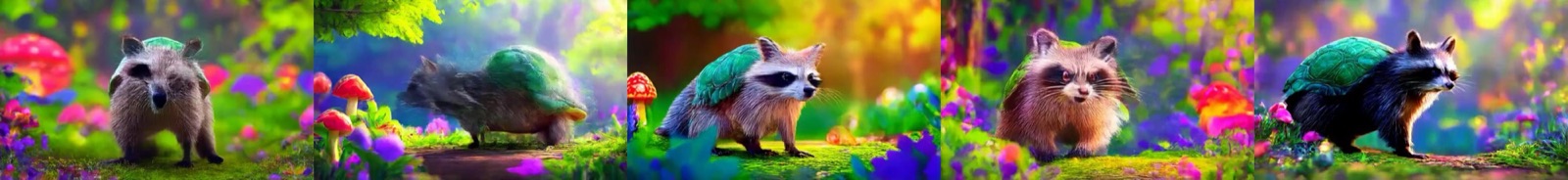} \\
\textbf{\duetgradient} & \includegraphics[width=\linewidth]{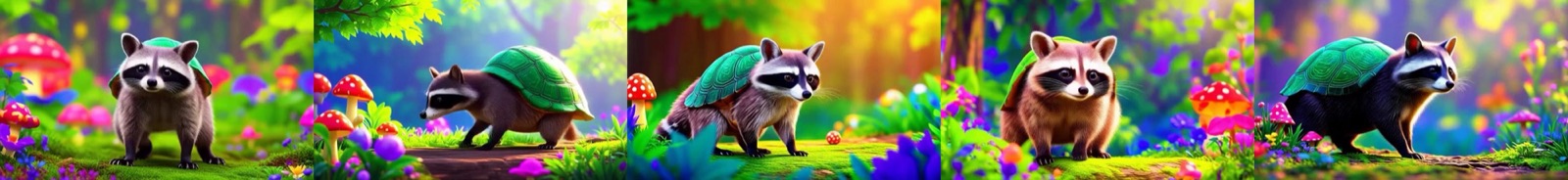} \\
\textbf{\duetplusgradient} & \includegraphics[width=\linewidth]{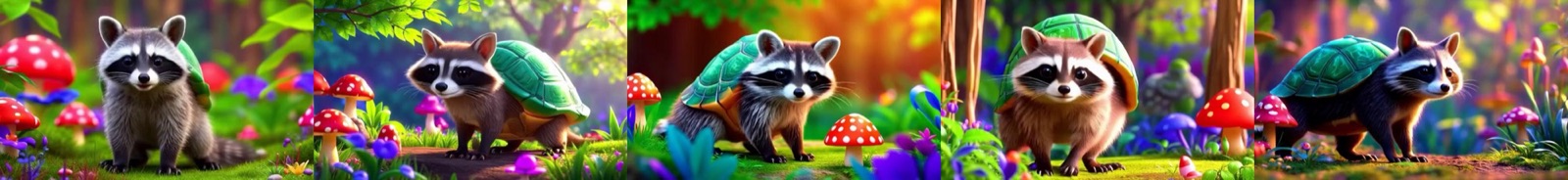} \\
\end{tabular}
\captionof{figure}{First frames of five generations on the prompt \textit{``A raccoon that looks like a turtle, digital art''}; each column is one shared seed.}
\label{fig:app-ff-raccoon}
\par}\vspace{6pt}

{\centering
\setlength{\tabcolsep}{2pt}
\begin{tabular}{@{}>{\centering\arraybackslash}m{0.13\textwidth}>{\centering\arraybackslash}m{0.72\textwidth}@{}}
& \appqualhdrseed \\
\textbf{DMD} & \includegraphics[width=\linewidth]{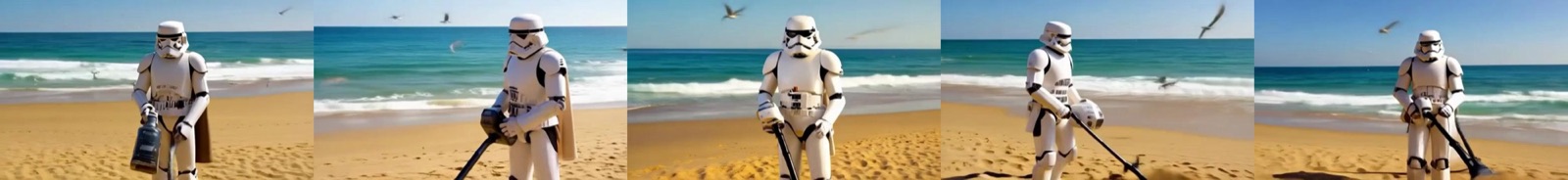} \\
\textbf{DP-DMD} & \includegraphics[width=\linewidth]{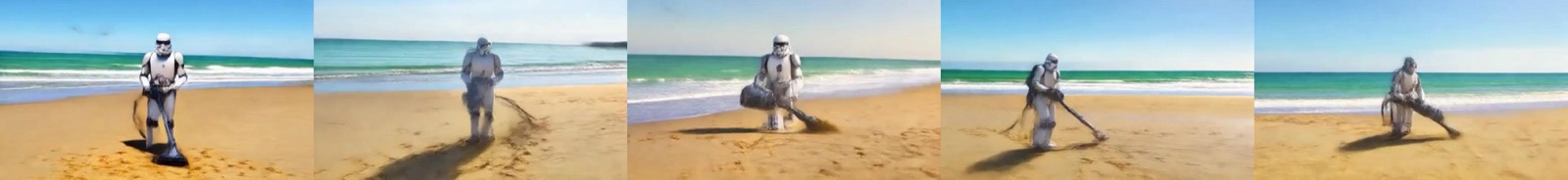} \\
\textbf{rCM} & \includegraphics[width=\linewidth]{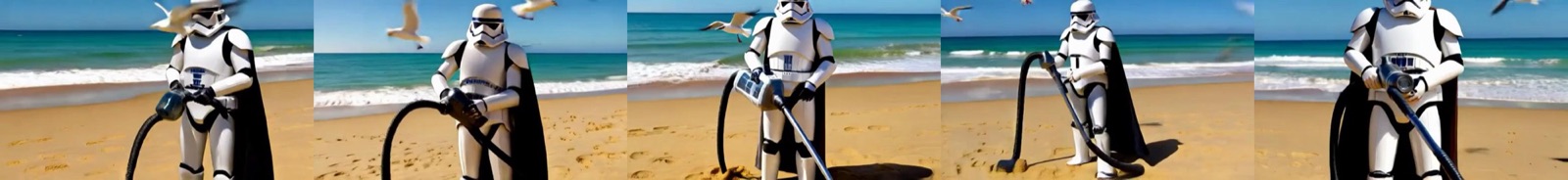} \\
\textbf{sCM} & \includegraphics[width=\linewidth]{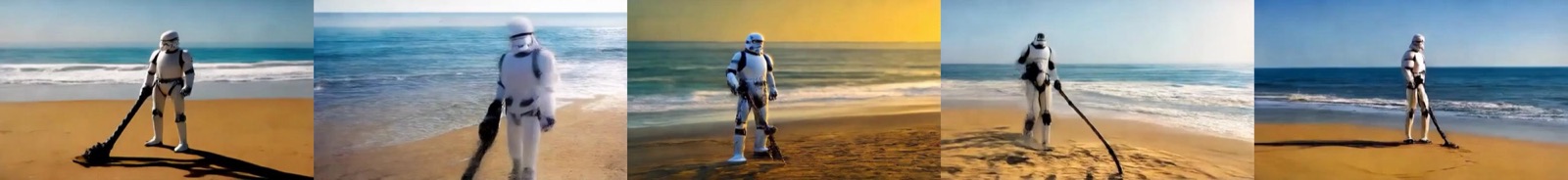} \\
\textbf{\duetgradient} & \includegraphics[width=\linewidth]{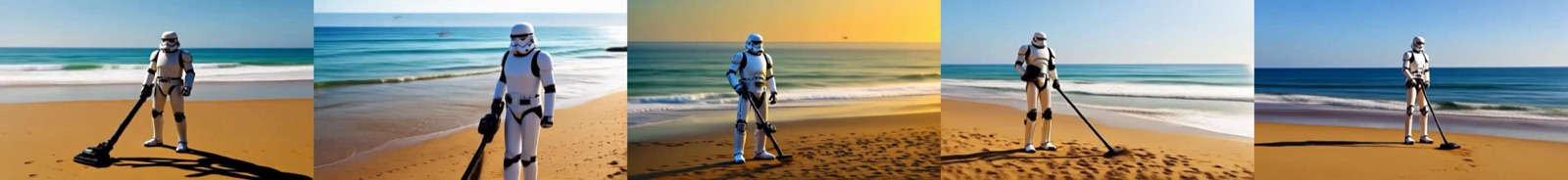} \\
\textbf{\duetplusgradient} & \includegraphics[width=\linewidth]{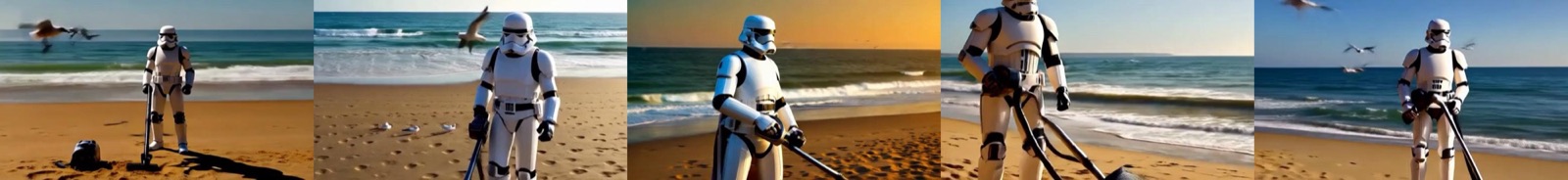} \\
\end{tabular}
\captionof{figure}{First frames of five generations on the prompt \textit{``A storm trooper vacuuming the beach''}; each column is one shared seed.}
\label{fig:app-ff-stormtrooper}
\par}\vspace{6pt}

{\centering
\setlength{\tabcolsep}{2pt}
\begin{tabular}{@{}>{\centering\arraybackslash}m{0.13\textwidth}>{\centering\arraybackslash}m{0.72\textwidth}@{}}
& \appqualhdrseed \\
\textbf{DMD} & \includegraphics[width=\linewidth]{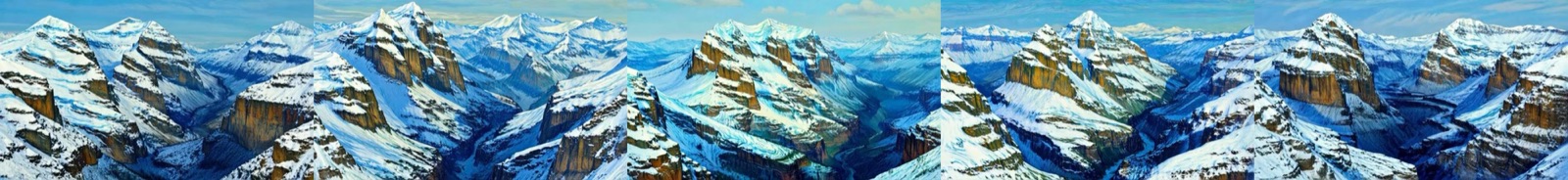} \\
\textbf{DP-DMD} & \includegraphics[width=\linewidth]{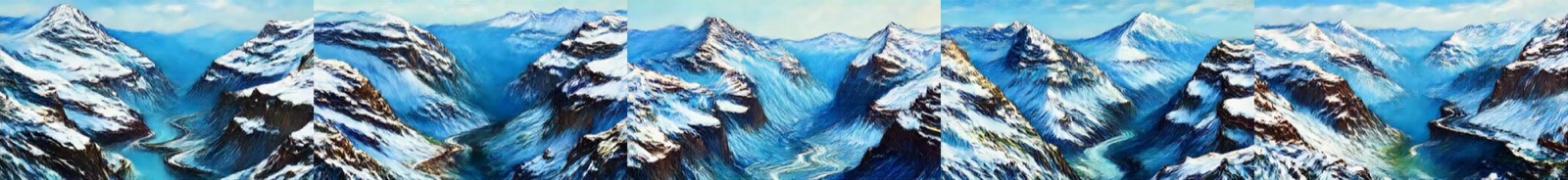} \\
\textbf{rCM} & \includegraphics[width=\linewidth]{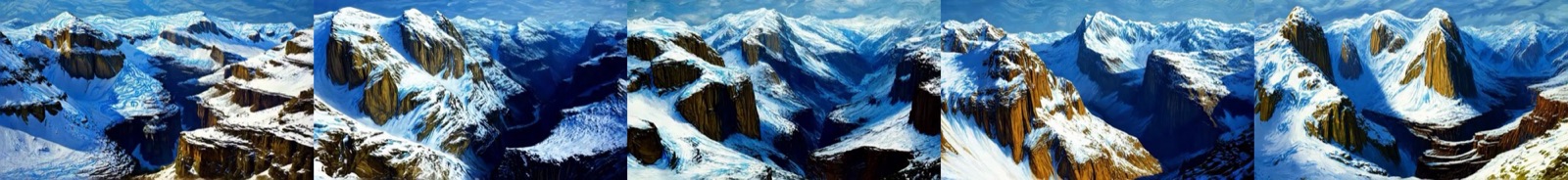} \\
\textbf{sCM} & \includegraphics[width=\linewidth]{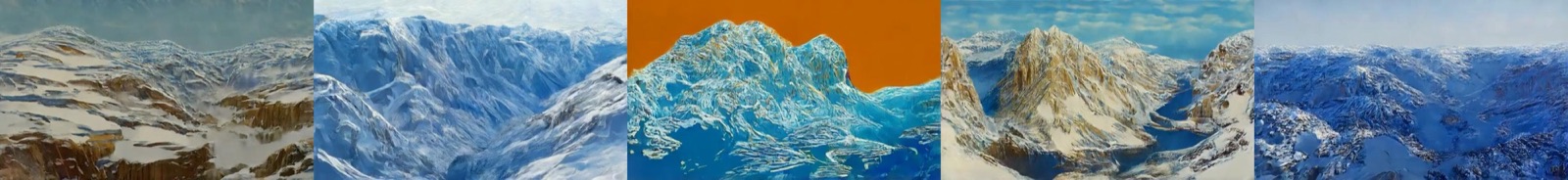} \\
\textbf{\duetgradient} & \includegraphics[width=\linewidth]{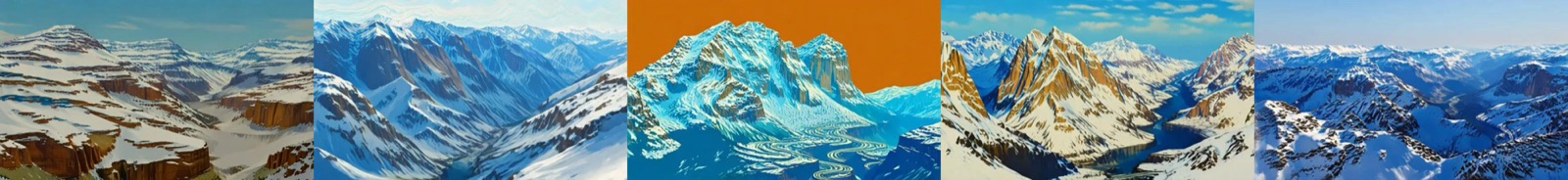} \\
\textbf{\duetplusgradient} & \includegraphics[width=\linewidth]{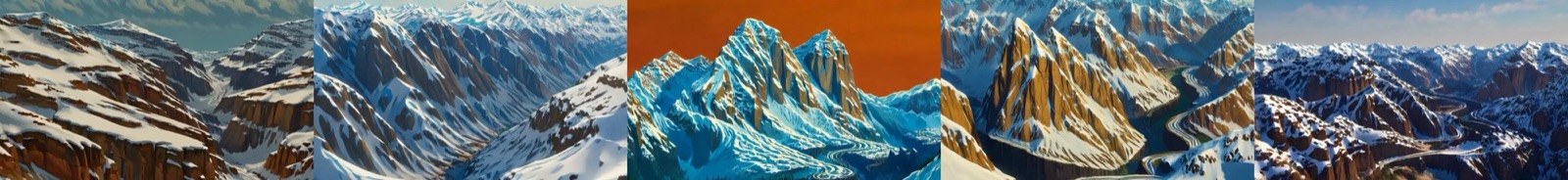} \\
\end{tabular}
\captionof{figure}{First frames of five generations on the prompt \textit{``Snow rocky mountains peaks canyon. snow blanketed rocky mountains surround and shadow deep canyons. the canyons twist and bend through the high elevated mountain peaks''}; each column is one shared seed.}
\label{fig:app-ff-snow}
\par}\vspace{6pt}

{\centering
\setlength{\tabcolsep}{2pt}
\begin{tabular}{@{}>{\centering\arraybackslash}m{0.13\textwidth}>{\centering\arraybackslash}m{0.72\textwidth}@{}}
& \appqualhdrframe \\
\textbf{DMD} & \includegraphics[width=\linewidth]{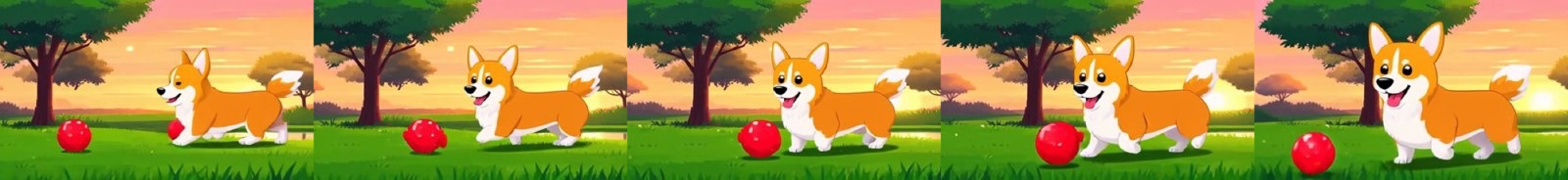} \\
\textbf{DP-DMD} & \includegraphics[width=\linewidth]{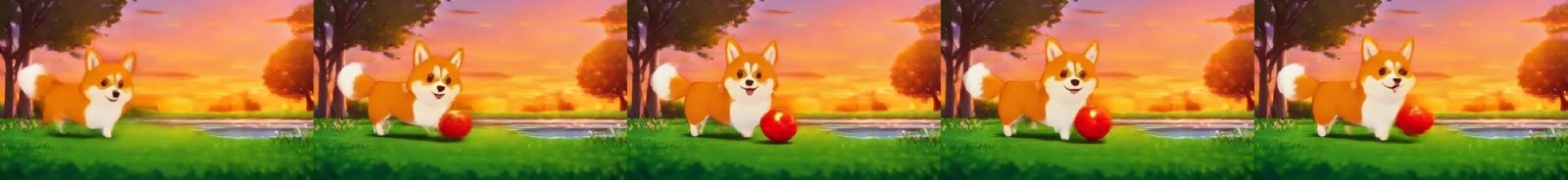} \\
\textbf{rCM} & \includegraphics[width=\linewidth]{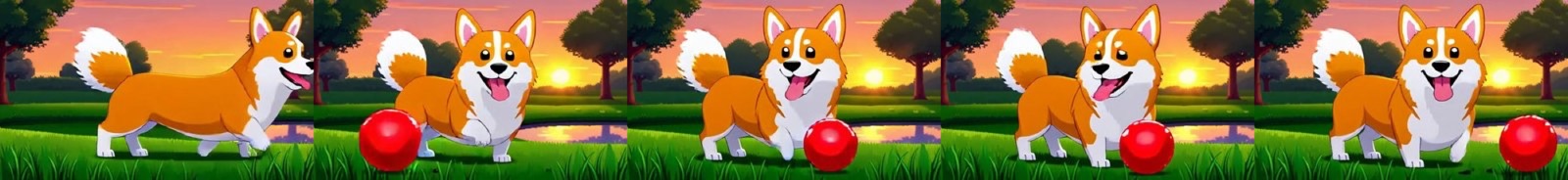} \\
\textbf{sCM} & \includegraphics[width=\linewidth]{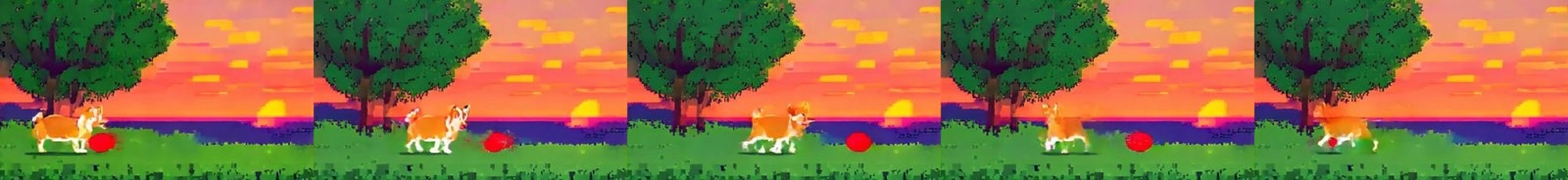} \\
\textbf{\duetgradient} & \includegraphics[width=\linewidth]{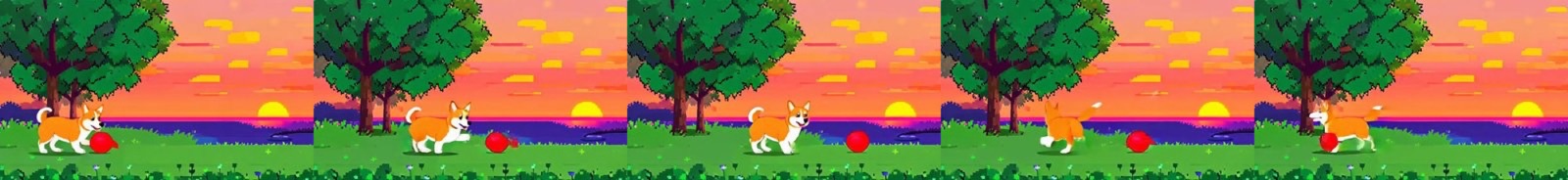} \\
\textbf{\duetplusgradient} & \includegraphics[width=\linewidth]{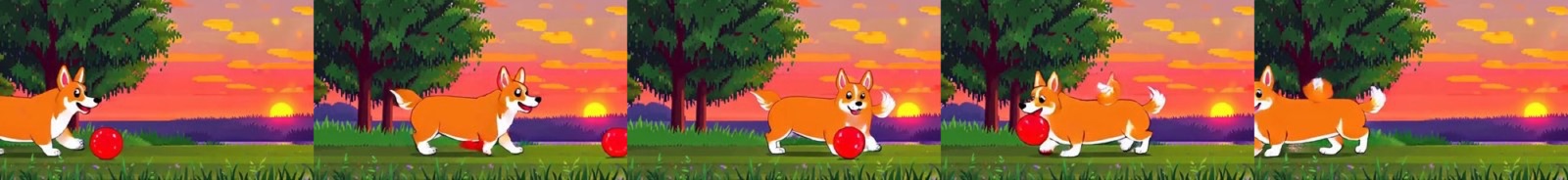} \\
\end{tabular}
\captionof{figure}{Five frames of generation \#4 on the prompt \textit{``A cute happy Corgi playing in park, sunset, pixel art''} (shared across models); each column is one time step.}
\label{fig:app-vf-corgi}
\par}\vspace{6pt}

{\centering
\setlength{\tabcolsep}{2pt}
\begin{tabular}{@{}>{\centering\arraybackslash}m{0.13\textwidth}>{\centering\arraybackslash}m{0.72\textwidth}@{}}
& \appqualhdrframe \\
\textbf{DMD} & \includegraphics[width=\linewidth]{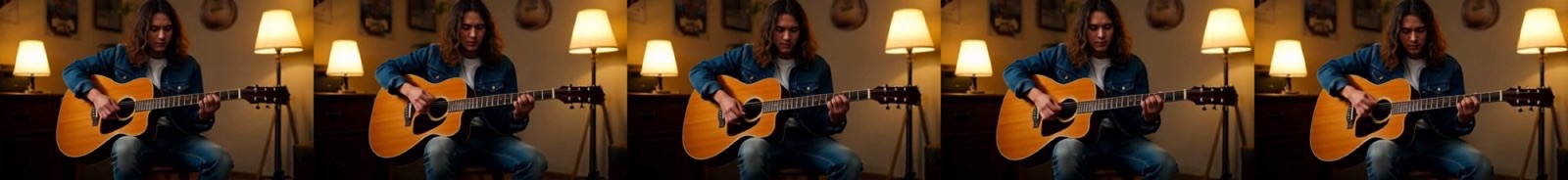} \\
\textbf{DP-DMD} & \includegraphics[width=\linewidth]{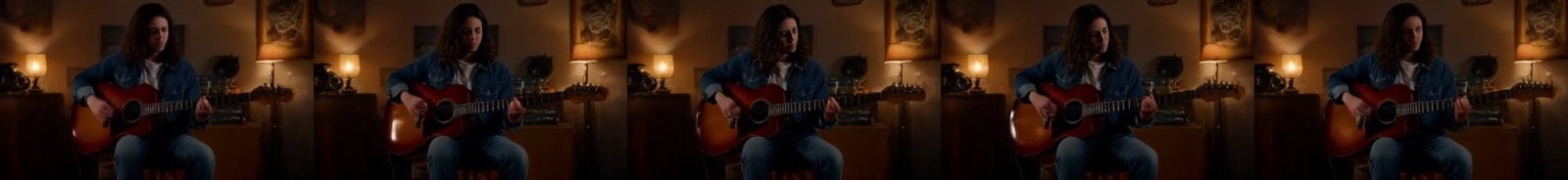} \\
\textbf{rCM} & \includegraphics[width=\linewidth]{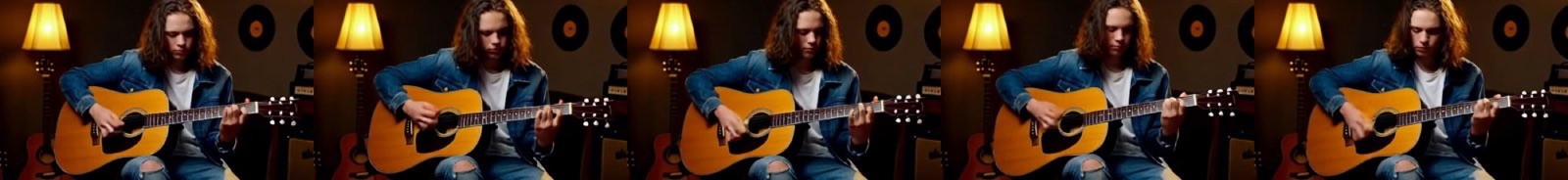} \\
\textbf{sCM} & \includegraphics[width=\linewidth]{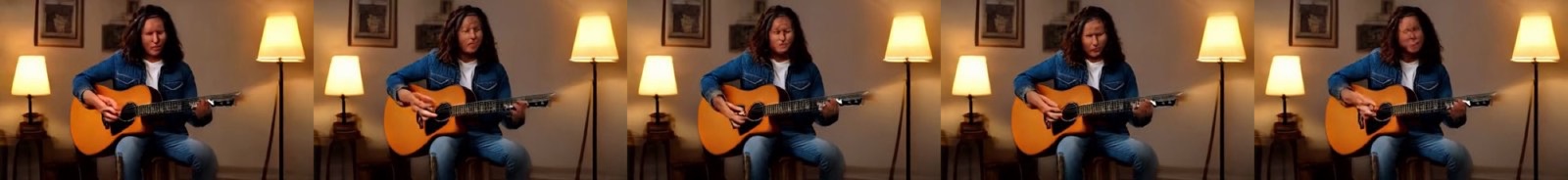} \\
\textbf{\duetgradient} & \includegraphics[width=\linewidth]{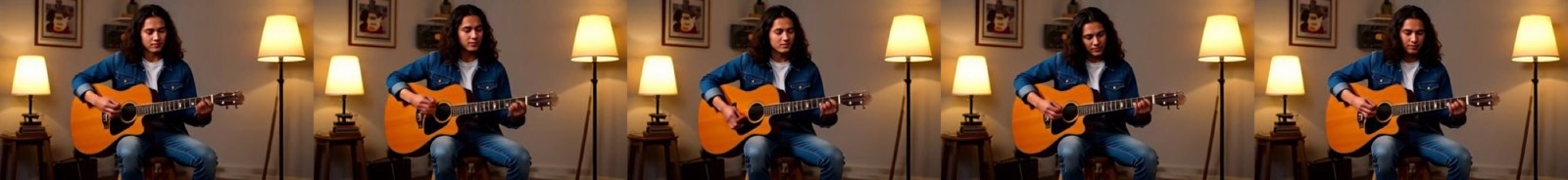} \\
\textbf{\duetplusgradient} & \includegraphics[width=\linewidth]{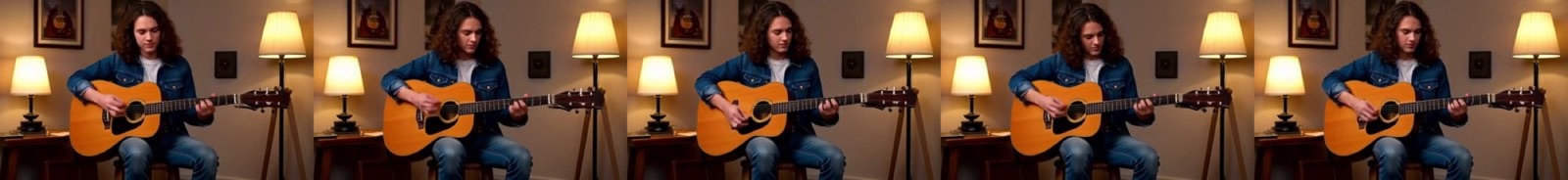} \\
\end{tabular}
\captionof{figure}{Five frames of generation \#1 on the prompt \textit{``A person playing guitar''} (shared across models); each column is one time step.}
\label{fig:app-vf-guitar}
\par}\vspace{6pt}

{\centering
\setlength{\tabcolsep}{2pt}
\begin{tabular}{@{}>{\centering\arraybackslash}m{0.13\textwidth}>{\centering\arraybackslash}m{0.72\textwidth}@{}}
& \appqualhdrframe \\
\textbf{DMD} & \includegraphics[width=\linewidth]{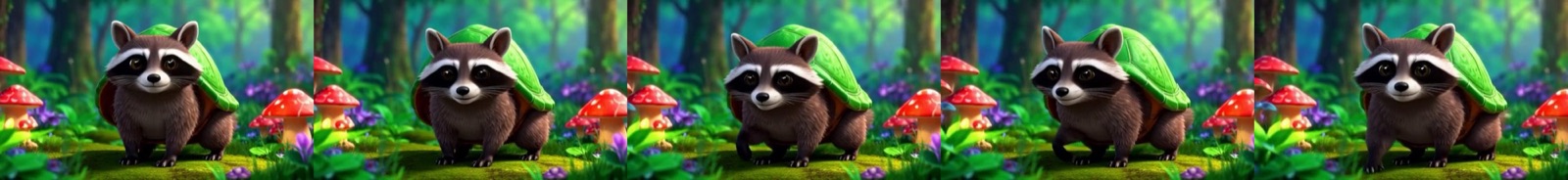} \\
\textbf{DP-DMD} & \includegraphics[width=\linewidth]{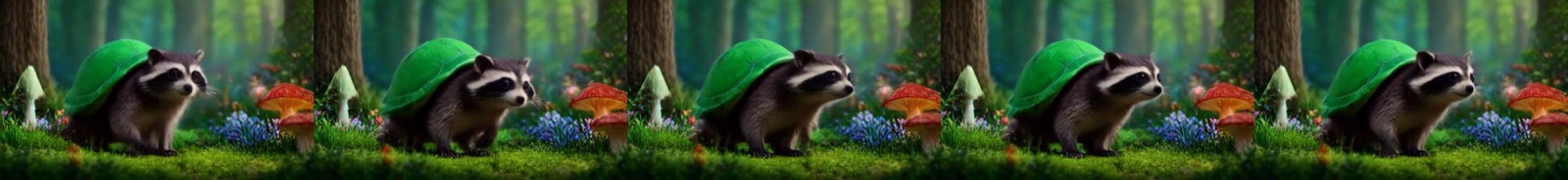} \\
\textbf{rCM} & \includegraphics[width=\linewidth]{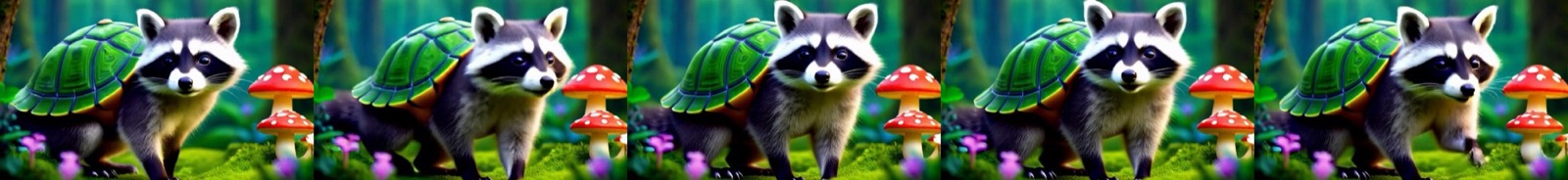} \\
\textbf{sCM} & \includegraphics[width=\linewidth]{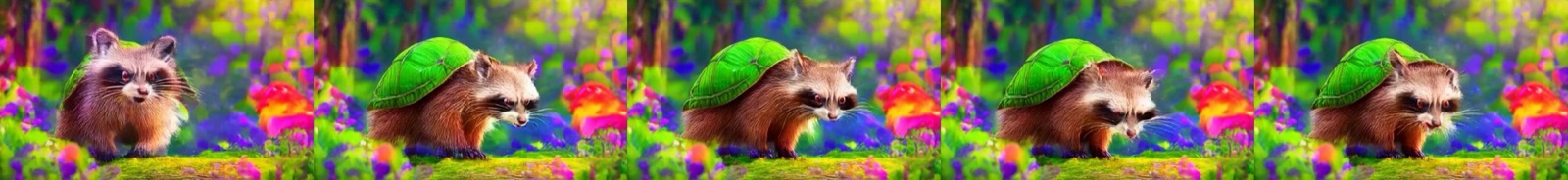} \\
\textbf{\duetgradient} & \includegraphics[width=\linewidth]{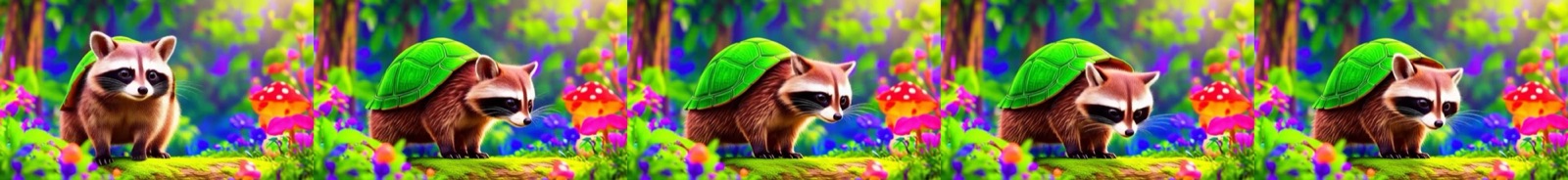} \\
\textbf{\duetplusgradient} & \includegraphics[width=\linewidth]{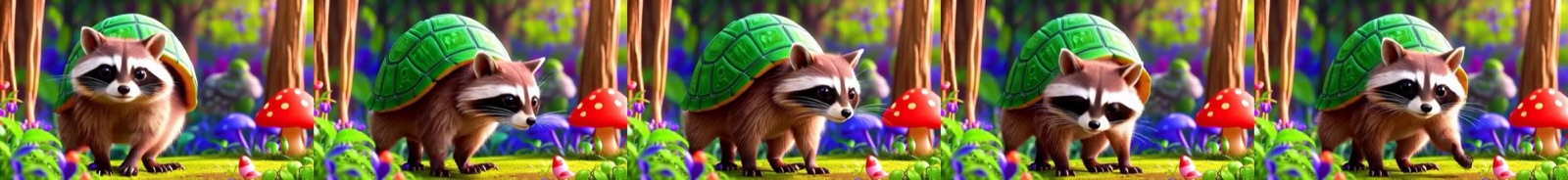} \\
\end{tabular}
\captionof{figure}{Five frames of generation \#4 on the prompt \textit{``A raccoon that looks like a turtle, digital art''} (shared across models); each column is one time step.}
\label{fig:app-vf-raccoon}
\par}\vspace{6pt}

{\centering
\setlength{\tabcolsep}{2pt}
\begin{tabular}{@{}>{\centering\arraybackslash}m{0.13\textwidth}>{\centering\arraybackslash}m{0.72\textwidth}@{}}
& \appqualhdrframe \\
\textbf{DMD} & \includegraphics[width=\linewidth]{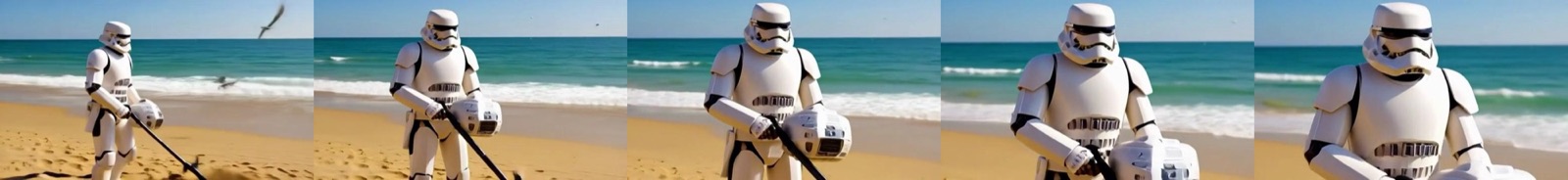} \\
\textbf{DP-DMD} & \includegraphics[width=\linewidth]{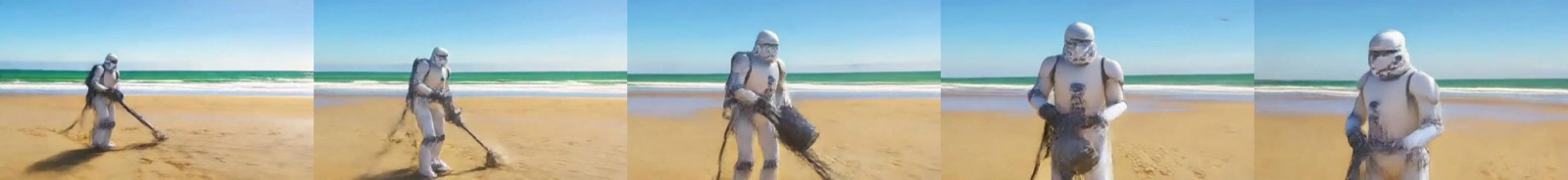} \\
\textbf{rCM} & \includegraphics[width=\linewidth]{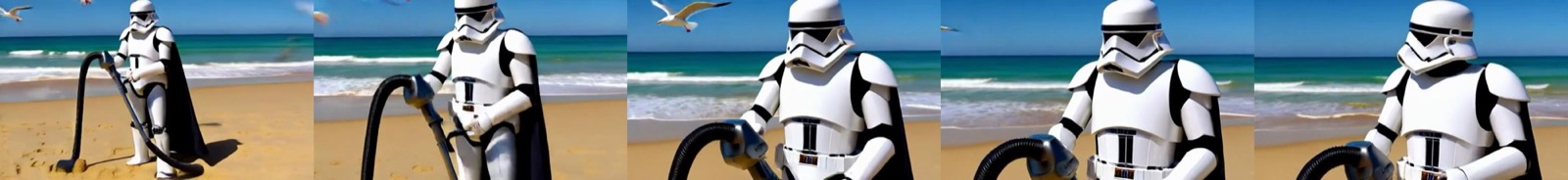} \\
\textbf{sCM} & \includegraphics[width=\linewidth]{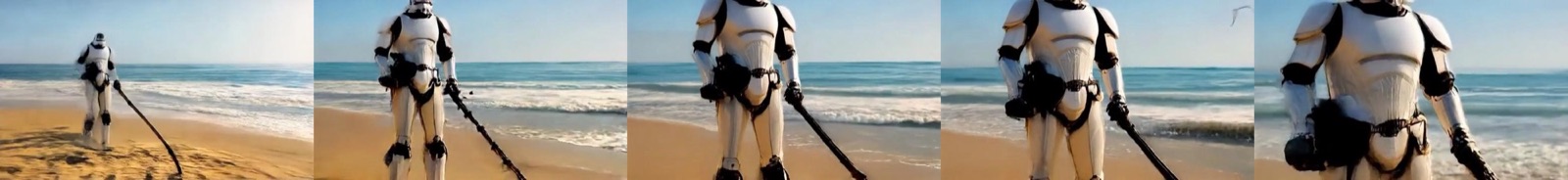} \\
\textbf{\duetgradient} & \includegraphics[width=\linewidth]{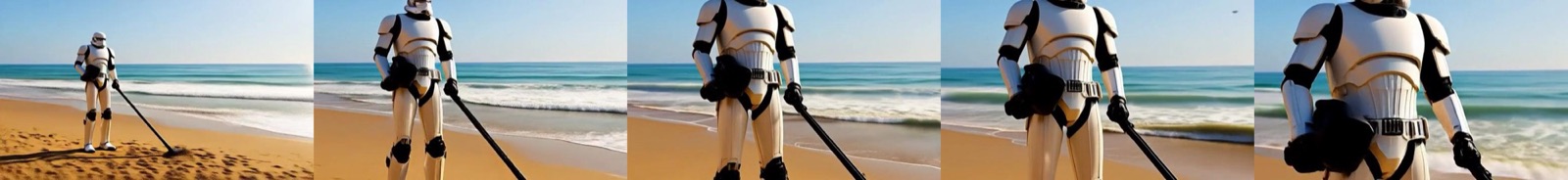} \\
\textbf{\duetplusgradient} & \includegraphics[width=\linewidth]{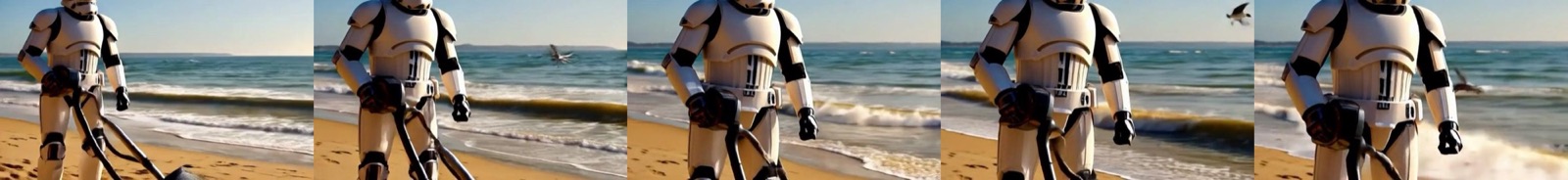} \\
\end{tabular}
\captionof{figure}{Five frames of generation \#4 on the prompt \textit{``A storm trooper vacuuming the beach''} (shared across models); each column is one time step.}
\label{fig:app-vf-stormtrooper}
\par}\vspace{6pt}

{\centering
\setlength{\tabcolsep}{2pt}
\begin{tabular}{@{}>{\centering\arraybackslash}m{0.13\textwidth}>{\centering\arraybackslash}m{0.72\textwidth}@{}}
& \appqualhdrframe \\
\textbf{DMD} & \includegraphics[width=\linewidth]{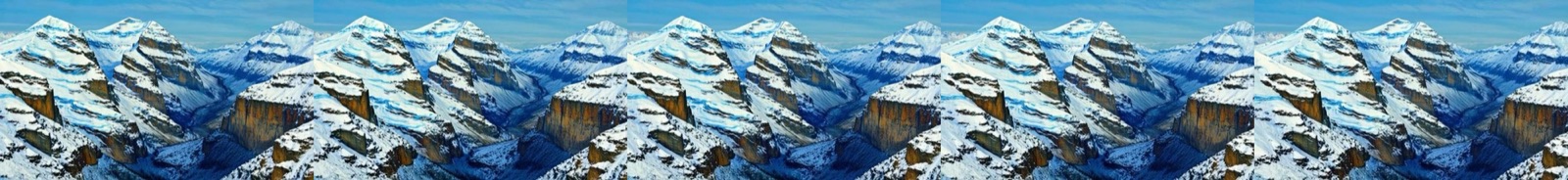} \\
\textbf{DP-DMD} & \includegraphics[width=\linewidth]{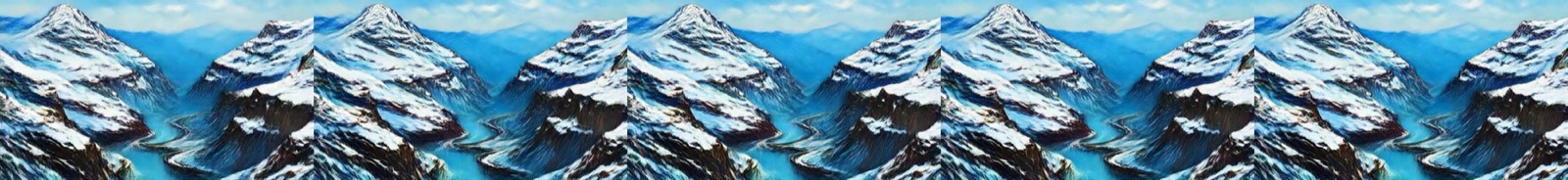} \\
\textbf{rCM} & \includegraphics[width=\linewidth]{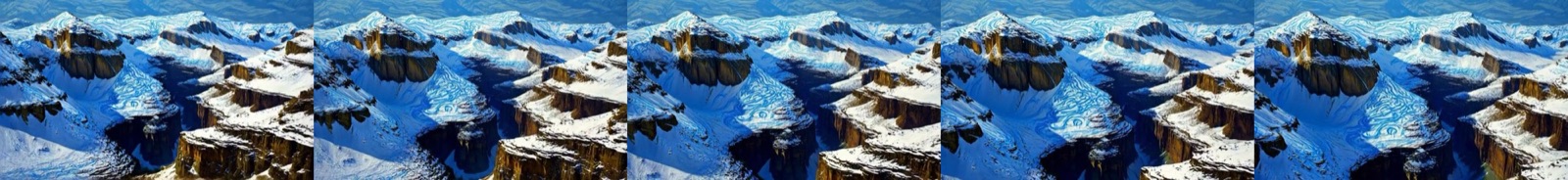} \\
\textbf{sCM} & \includegraphics[width=\linewidth]{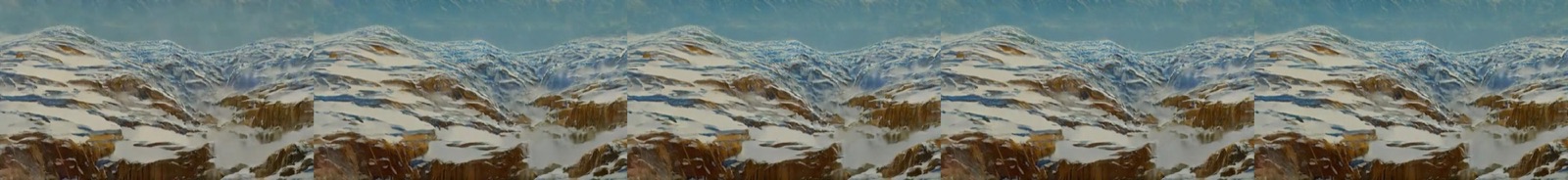} \\
\textbf{\duetgradient} & \includegraphics[width=\linewidth]{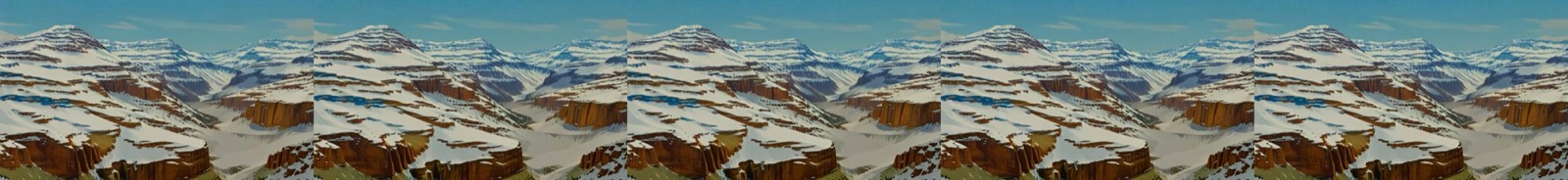} \\
\textbf{\duetplusgradient} & \includegraphics[width=\linewidth]{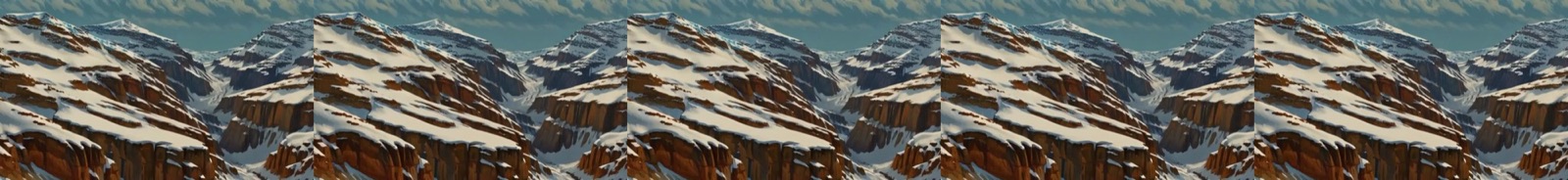} \\
\end{tabular}
\captionof{figure}{Five frames of generation \#1 on the prompt \textit{``Snow rocky mountains peaks canyon. snow blanketed rocky mountains surround and shadow deep canyons. the canyons twist and bend through the high elevated mountain peaks''} (shared across models); each column is one time step.}
\label{fig:app-vf-snow}
\par}\vspace{6pt}

\end{document}